\documentclass[pdflatex,sn-basic]{sn-jnl}

\usepackage{graphicx}
\usepackage{multirow}
\usepackage{amsmath,amssymb,amsfonts}
\usepackage{mathtools}
\usepackage{bm}
\usepackage{xcolor}
\usepackage{booktabs}
\usepackage{tabularx}
\usepackage{threeparttable}
\usepackage{tablefootnote}
\usepackage{float}
\usepackage{placeins}
\usepackage[normalem]{ulem}
\usepackage[title]{appendix}
\usepackage{comment}
\usepackage{enumerate}
\usepackage{hyperref}

\theoremstyle{thmstyleone}

\theoremstyle{thmstyletwo}

\theoremstyle{thmstylethree}

\makeatletter
\providecommand{\footnoteA}[1]{%
  \begingroup
  \renewcommand\thefootnote{}%
  \renewcommand\@makefnmark{}%
  \footnotetext{#1}%
  \endgroup}
\makeatother

\begin{document}

\title{Physics-Informed Neural Embeddings of PDE Solution Families}

\author[1,2]{\fnm{Raul} \sur{Jimenez}}
\author[3]{\fnm{Svitlana} \sur{Mayboroda}}
\author[4]{\fnm{Pavlos} \sur{Protopapas}}
\author*[1,5]{\fnm{Leonid} \sur{Sarieddine}}\email{leonid.sarieddine@icc.ub.edu}
\author[6]{\fnm{David N.} \sur{Spergel}}
\author*[1,5]{\fnm{Pedro} \sur{Taranc\'on-\'Alvarez}}\email{pedro.tarancon@icc.ub.edu}

\affil*[1]{\orgdiv{Institute of Cosmos Sciences (ICC)}, \orgname{University of Barcelona}, \orgaddress{\street{Mart\'i i Franqu\`es 1}, \postcode{ES-08028}, \city{Barcelona}, \country{Spain}}}
\affil[2]{\orgname{ICREA}, \orgaddress{\street{Pg. Llu\'is Companys 23}, \postcode{08010}, \city{Barcelona}, \country{Spain}}}
\affil[3]{\orgdiv{Department of Mathematics}, \orgname{ETH Zurich}, \orgaddress{\street{R\"amistrasse 101}, \postcode{8092}, \city{Z\"urich}, \country{Switzerland}}}
\affil[4]{\orgdiv{Institute for Applied Computational Science}, \orgname{Harvard University}, \orgaddress{\city{Cambridge}, \state{MA}, \country{USA}}}
\affil[5]{\orgdiv{Department of F\'isica Qu\`antica i Astrof\'isica}, \orgname{Universitat de Barcelona}, \orgaddress{\street{Mart\'i i Franqu\`es 1}, \postcode{ES-08028}, \city{Barcelona}, \country{Spain}}}
\affil[6]{\orgname{Flatiron Institute}, \orgaddress{\street{162 Fifth Avenue}, \city{New York}, \state{NY}, \postcode{10011}, \country{USA}}}

\abstract{We introduce a physics-informed framework for learning finite-dimensional embeddings of solution families of partial differential equations. The method uses a multihead Physics-Informed Neural Network in which a shared body learns a latent manifold representing the solution space, while linear heads reconstruct individual solutions associated with different initial conditions. A head-orthogonalization penalty removes degeneracies in the latent representation and stabilizes the principal-component spectrum across training realizations. Because the initial condition is built into the network output by construction, these principal components measure the additional variability the network learns on top of the initial profile, not the full solution itself. We apply the method to the one-dimensional viscous Burgers equation, with the heat and wave equations as robustness checks. For a latent dimension $n_b=20$, the learned manifolds exhibit pronounced effective dimensional reduction: for Burgers dynamics, only $2$--$4$ principal components capture about $95\%$ of the latent-space variance, while $4$--$7$ capture about $99\%$, depending on the initial-condition family; the same qualitative compression holds for the heat and wave equations. We also split the wavenumber axis into bands (``Fourier shells'') and measure how much each band contributes to every principal component. The resulting frequency profile is invariant under the change-of-basis freedom that the orthogonalization penalty leaves in the latent space, and is therefore reproducible across independent training runs. More broadly, this establishes the learned spectral profiles and principal components as robust observables of solution-manifold geometry.}

\keywords{Physics-Informed Neural Networks, partial differential equations, latent representations, multihead neural networks, reduced-order modelling, Burgers equation, solution manifolds}

\artnote{The authors are listed in alphabetical order.}

\maketitle

\section{Introduction}\label{sec: introduction}

A central question in physics-informed machine learning is whether the space of solutions of a nonlinear partial differential equation (PDE) can be understood as a low-dimensional geometric object, and whether neural networks (NNs) can be used to uncover the geometry directly from the governing equations. This question is physically motivated: nonlinear PDEs generate solution families with intricate structure arising from interactions across scales, from shock formation in viscous flows to multiscale transport in turbulence (\cite{Frisch1995, Doering1995, AlexakisBiferale2018}). The Navier--Stokes equations stand as a paradigmatic example, but the question applies broadly to any system in which a compact description of the solution manifold would provide interpretable insight beyond individual numerical realizations. The mathematical complexity of these equations --- global regularity in 3D remains an open Millennium Problem (\cite{Fefferman2000}), and non-uniqueness of weak solutions has only recently been established (\cite{BuckmasterVicol2019}) --- underscores that solution spaces of nonlinear PDEs have structure that is far from fully understood, even analytically.

Traditionally, progress in this area has relied on a combination of analytical approximations, phenomenological modeling, and increasingly sophisticated numerical simulations. While these approaches have achieved remarkable success in reproducing observed phenomena, numerical solvers produce individual realizations corresponding to specific initial or boundary conditions (IC/BC) and do not directly expose the structure of the solution space. Classical reduced-order techniques --- proper orthogonal decomposition (\cite{Berkooz1993}), reduced-basis methods (\cite{quarteroni2015reduced}), and Koopman-based approaches such as dynamic mode decomposition (\cite{Schmid2010, Mezic2005}) --- partially address this gap, but they work post-hoc on precomputed snapshots and are not coupled to the governing equations during basis construction. As a result, the basis functions they produce reflect the statistics of the precomputed snapshot ensemble rather than a representation informed by the governing equations for the chosen family of solutions. This limitation becomes especially acute in high-dimensional or multiscale systems, where the sheer volume of data generated by simulations can obscure rather than clarify the essential physics.

In parallel with these developments, recent advances in machine learning (ML) have introduced a complementary paradigm based on representation learning. In this framework, high-dimensional data are projected into a more semantically meaningful embedding --- typically of lower dimension, though sometimes higher --- in which their essential structure becomes more interpretable (\cite{Bengio2013, Goodfellow2016}). This raises the question of whether similar ideas can be applied to physical systems governed by PDEs: namely, whether the space of solutions of a nonlinear equation can itself be understood as a geometric object embedded in a lower-dimensional representation space. Addressing this question requires methods capable of learning representations that remain constrained by the governing equations themselves, rather than solely by collections of precomputed solutions.

Physics-Informed Neural Networks (PINNs) provide a natural framework for this purpose precisely because they incorporate the governing differential equations (DEs) directly into the training objective, allowing NNs to approximate solutions while enforcing physical consistency (\cite{raissi2019, mattheakis2019symmetries, Karniadakis2021}). Unlike traditional solvers, which must be rerun for each choice of parameters or ICs, PINNs can be trained to represent parametric families of solutions by treating parameters, BCs, ICs, and forcing terms as inputs to the network (\cite{Flamant2020, Cuomo2022, lu2021deepxde}). In this sense, PINNs define an implicit mapping from a parameter space to a space of functions, thereby providing a representation of the solution manifold associated with the PDE. Crucially, this representation is learned under the governing equation and BC/IC constraints during training, rather than constructed post-hoc from precomputed solution datasets. Related equation-informed approaches for constructing reduced representations of parametric PDEs have also been developed within reduced-basis and manifold-approximation frameworks (\cite{lassila2013generalized, quarteroni2015reduced}).

In parallel with the development of PINNs, a closely related line of research has focused on \emph{operator learning}, in which the goal is to approximate mappings between infinite-dimensional function spaces rather than individual solutions. Prominent examples include Deep Operator Networks (DeepONet) (\cite{Lu2021DeepONet}) and Fourier Neural Operators (FNO) (\cite{Li2021FNO, Li2023FNO}), which learn solution operators that map initial or boundary data directly to the corresponding solutions of a PDE. These approaches have demonstrated remarkable efficiency and generalization capabilities, particularly in parametric and high-dimensional settings, and have recently been extended to incorporate physical constraints through hybrid architectures such as physics-informed neural operators (\cite{Goswami2022PINO}). From this perspective, the solution manifold of a PDE is implicitly encoded as the image of a learned operator acting on a space of inputs. 

In contrast, the present work does not learn a solution operator, but instead constructs a finite-dimensional representation of the solution family via latent coordinates and learned basis functions. To this aim, we use a multihead (MH) PINN to learn an explicit, finite-dimensional embedding of the PDE solution manifold. Each solution is represented through low-dimensional coordinates together with a set of learned basis functions, so that the network directly parametrizes a compact representation of the solution family. A head-orthogonalization penalty removes degeneracies in the latent representation and stabilizes the principal-component (PC) spectrum across training realizations, making it a robust, training-independent observable. We further split the wavenumber axis into bands (``Fourier shells'') and measure how much each band contributes to every PC, giving each PC a frequency profile that is invariant under the change-of-basis freedom that the orthogonalization penalty leaves in the latent space and is therefore reproducible across independent training runs. This provides a concrete diagnostic of how information at different physical scales is organized within the learned solution manifold, going beyond variance-based dimensional reduction to a scale-resolved description of the embedding geometry. We do not aim to train a generalizable model that predicts solutions for unseen ICs outside the training set. The focus is on extracting training-independent geometric information from the latent space learned for a particular family of ICs.

As a controlled testbed, we focus on the one-dimensional viscous Burgers equation, which provides a minimal model for nonlinear advection and diffusion while retaining key qualitative features of more complex systems, including shock formation, dissipation, and energy transfer across scales (\cite{Burgers1948, Hopf1950}). It can be viewed as a simplified limit of the Navier--Stokes equations, making it an ideal setting in which to develop and validate new methodologies before extending them to higher-dimensional systems. In particular, it allows us to systematically study how the solution depends on initial conditions and viscosity under full numerical and analytical control.

Applying the method to this system, we find that only 2--4 principal components are required to capture approximately $95\%$ of the latent-space variance, despite a nominal embedding dimension of 20 --- indicating a pronounced effective dimensional reduction. This reveals a hierarchical organization of the latent space, in which a small number of dominant components account for most of the variance, with rapidly decreasing contributions from higher-index components. We also apply the method to the heat and wave equations as robustness checks under both dissipative and non-dissipative linear dynamics, allowing us to isolate which features of the learned embedding are specific to the nonlinearity of Burgers versus generic to the PINN framework. The same compression persists across qualitatively different IC families.

More broadly, these results support the view that meaningful geometric structure can emerge in representation spaces learned under explicit PDE constraints, even in the absence of precomputed solution datasets. This suggests that physics-informed representation learning can provide a direct route to extracting interpretable low-dimensional structure from the solution spaces of nonlinear PDEs. These findings open a concrete path toward equation-informed reduced-order modeling and toward the geometric analysis of nonlinear dynamical systems.

The paper is organized as follows. In Sec. \ref{sec: theory} we review the Burgers equation and its relation to the Navier--Stokes system. In Sec. \ref{sec: methodology} we introduce the PINN framework and describe the MH architecture used to learn the embedding space, together with the novel techniques (head orthogonalization and spectral decomposition of the latent space) developed in this work to extract degeneracy-free information from the embedding space. In Sec. \ref{sec: results} we present the numerical results for Burgers equation and explore the structure of the latent representations using principal component analysis (PCA) and spectral decomposition of the PCs. Finally, in Sec. \ref{sec: conclusions} we discuss the implications of our findings and outline directions for future work. We also include in this work four appendices. In appendix \ref{appendix: NN} we summarize the technical details of the NN architecture and the training process. In appendix \ref{appendix: IC} we explain how different IC families are generated. In appendix \ref{Appendix 1} we apply the setup to both the heat and wave equations. In appendix \ref{Appendix2} we give an expression for the error bound on the robustness of the PCA decomposition of the latent space depending on the loss function.

\section{Burgers equation and its relation to the Navier-Stokes system}\label{sec: theory}

The classic Navier--Stokes equations are among the central systems of hydrodynamics. They describe the dynamics of a non-relativistic viscous fluid. The equation can be written in the following form (\cite{Landau1987})
\begin{equation}\label{eq: Navier Stokes}
    \frac{\partial \vec{u}}{\partial t} + (\vec{u}\cdot \nabla)\vec{u} \,=\, -\frac{\nabla p}{\rho} + \nu \nabla^2 \vec{u} + \left(\xi +\frac{1}{3} \nu\right)\nabla \left(\nabla \cdot \vec{u}\right)
\end{equation}
where $\vec{u}$ represents the velocity of the fluid, $\rho$ the density, $p$ the pressure and $\nu$ and $\xi$ the shear and bulk kinematic viscosities respectively. Throughout this paper, we focus on the one-dimensional case. To obtain a closed scalar evolution equation, one typically replaces the pressure contribution by a phenomenological closure or assumes a regime in which pressure gradients can be absorbed into an external forcing or are dynamically negligible. Under this modeling assumption, the dynamics reduce to the viscous Burgers equation
\begin{equation}\label{eq: burgers equation}
    \frac{\partial u}{\partial t} + u\frac{\partial u}{\partial x}
    = \nu \frac{\partial^2 u}{\partial x^2}.
\end{equation}
The viscous Burgers equation can be viewed as a prototypical scalar conservation law with viscosity, sharing the same nonlinear advection--diffusion structure as Navier--Stokes but without the nonlocal pressure constraint. For a more detailed derivation, we refer the reader to Refs. \cite{Landau1987, Enflo2004}.

It is often used as a toy model for qualitative features of turbulent flow. The behavior depends strongly on the viscosity $\nu$. For positive viscosity ($\nu > 0$), one obtains the viscous Burgers equation. In this regime, there is no true turbulence in the classical sense, but the dynamics still show nonlinear effects such as shock formation regularized by diffusion and energy dissipation. For zero viscosity ($\nu = 0$), one obtains the inviscid Burgers equation. In this case, smooth initial conditions can develop shocks in finite time. The system remains much simpler than fully developed turbulence, but it exhibits features such as nonlinear transfer across scales without dissipation. We restrict attention to the viscous case in this work, while the inviscid limit is left for future study.

\section{Methodology}\label{sec: methodology}

This section presents the methodological framework for learning and analyzing low-dimensional embeddings of PDE solution families. We build on PINNs and extend them to MH architectures in which a shared body learns a latent representation of the solution manifold while linear heads map this embedding to solutions associated with different ICs/BCs. We further introduce a head-orthogonalization penalty to remove degeneracies in the principal components of the latent space and define a spectral decomposition of the latent covariance that quantifies scale-resolved structure across different initial-condition families.

\subsection{Physics-Informed Neural Networks background}\label{subsec: PINNs}

PINNs, first introduced in the works of \cite{Dissanayake1994} and \cite{Lagaris2000}, have emerged as a powerful tool for solving PDEs and ODEs. \cite{raissi2019} proved this approach to be valid on a wide variety of challenging physical problems, while other researchers have made significant advancements in applying NNs to PDEs and ODEs, including notable examples such as \cite{Cai2019,Mattheakis2020,Sirignano2018}. In this paradigm, one typically trains a distinct NN for each unknown function in the governing equations. The training process involves minimizing a loss function that encodes the squared residuals of the DEs, thereby embedding the physics directly into the learning objective.

Throughout this work, we consider standard feedforward fully connected neural networks (FCNNs) with element-wise nonlinear activation functions. Details of the architecture are given in appendix \ref{appendix: NN}.

In the standard PINNs construction, we use one NN for every unknown function appearing in the DEs. We denote these objects as $\psi^{NN}(x^\mu)$, which represent the approximate solutions produced by the NNs, evaluated at different points of the independent variables and, when applicable, for different parameter values or IC/BCs, all of them denoted by $x^\mu$, which represents the inputs to the model\footnote{Note that $x^\mu$ includes both spatiotemporal coordinates and different parameters within the same DE form. The distinction between these two will be made precise in the following sections.}. The training process consists of minimizing a loss function, typically defined as the sum of squared residuals of the DEs evaluated over all sample points in the input domain:
\begin{equation}\label{eq: loss function}
    L \,=\, \sum_{\mu = \text{batch}}(D\psi^{NN}(x^\mu) - f(x^\mu))^2 + \lambda L_{add}
\end{equation}
where $D$ is a differential operator and $f(x^\mu)$ a source/forcing term, both depending on the particular DE (with $f \equiv 0$ for the homogeneous equations considered in this work). The term $L_{\mathrm{add}}$ represents additional contributions to the loss function that can be included to provide extra information during training. Its relative weight is controlled by the hyperparameter $\lambda$.

While traditional numerical methods often outperform PINNs in terms of computational efficiency and accuracy, recent developments in PINNs are offering unique advantages. One key benefit is the ability of PINNs to learn parametric dependence of the solutions to different ICs/BCs, or even different values of parameters appearing in the equations (\cite{Flamant2020,Cuomo2022,Karniadakis2021,lu2021deepxde}). In contrast with traditional numerical methods, this allows us to train the model on a wide range of different parameters at once, whereas a traditional solver must be rerun for different values of ICs/BCs, or parameters.

Another key advantage of PINNs is their capability of learning a general, high-dimensional space of functions called the \textit{latent} or \textit{embedding space}. The final solution is simply a combination of these functions in a certain way. This latent space depends, in general, on the independent variables of the problem, but it can also be learned for different ICs/BCs or different values of parameters appearing in the equations (\cite{Flamant2020}). In this work, we learn a latent space of Burgers equation for different values of the viscosity $\nu$, and different ICs in the so-called multihead (MH) training \citep{pellegrin2022transfer, desai2022oneshot}. We also learn this embedding space for the heat and wave equations, for different values of the heat diffusivity and the wave speed respectively. We will explain the proposed framework in the following subsection.

PINNs offer further advantages: PINNs are mesh-free, allowing one to obtain a solution at any point in the domain. Moreover, they can also be used to find new solutions on stiff regimes via transfer learning (\cite{auroy2025one,tarancon2025efficient}). PINNs have also been widely used to tackle inverse problems (\cite{raissi2019, yang2021bayesian, Bea:2024xgv, tarancon2025efficient}) due to their ability to learn the dependence of DEs solutions on boundary data.

\subsection{Multihead architecture}\label{subsec: MH}
In this section, we describe one of the key components used throughout this work: the MH architecture and training. This consists of splitting the NN $\psi^{NN}$ into two parts: the body, which learns the latent or embedding space; and the head, which maps the latent components to the final solution. This architecture has already been shown to be useful for transfer learning of DE solutions in stiff regimes \citep{tarancon2025efficient}. Here, we instead use MH training to analyze the mathematical structure of the embedding space, providing insight into the functional basis learned by the model to represent the solution.

\subsubsection{Basic concepts}

The MH approach for solving DEs with PINNs aims to learn the solution manifold (latent or embedding space) of a given system, rather than a single specific solution. This formulation allows the model to account for variations in ICs, BCs, system parameters, and forcing terms $f$ in the inhomogeneous part of the DE. By capturing the dependence of the solution on these factors, the model can generalize across a broad family of DE problems.

To formalize this approach, we represent the DE system as follows:
\begin{equation}
    D\psi^{NN}(x^\mu) - f(x^\mu) = 0\qquad ; \qquad\mu = 1, 2, 3, \ldots
\end{equation}
where each $x^\mu$ represents different inputs to the model. $D$ is a generic differential operator that encodes the form of the DE system, $\{x^{\mu}\}_{\mu = 1,.., m-1}$ are the independent variables, $\{x^\mu\}_{\mu = m,.., d}$ is a list of parameters describing the solutions for different ICs, BCs or parameters of the equations, and $\psi^{NN}(x^\mu)$ is the solution of the DE system. $f(x^\mu)$ is a general function that depends on the independent variable and may also depend on a parameter.

From this point forward, we will refer to any scenario with the same DE form (same $D$) and some choice of ICs, BCs, parameters and/or forcing functions $f$ as belonging to the same \textit{family} of DEs. Any variation of $\{x^\mu\}_{\mu = m,.., d}$ and/or $f$ will be a variation within the same family of the DEs, and a specific choice of them will be an \textit{element} of the family of DEs.

The MH approach is based on learning a mapping to an embedding space that generalizes across different solutions within the same family of DEs. This space is represented by a set of $n_b$ functions, $H_i(x^\mu)$, which are generated by the \textit{body} NN. This NN contains a set of weights and biases that we denote by $W_B$.

The latent space of solutions can then generate a specific solution of the family of the DE system by combining the $n_b$ components. This mapping is performed by another NN, known as the \textit{head}. The NN produces an approximate solution $\psi^{NN}(x^\mu)$ of a DE. The final result of the head can be expressed as follows 
\begin{equation*}
    \psi^{NN}(x^\mu) \,=\, \text{head}_{W_j}[H_i(x^\mu)]
\end{equation*}
where $\text{head}$ denotes the response of the head, and $W_j$ the set of weights and biases of the $j$-th head model.

\begin{figure*}[t]
  \centering
  \includegraphics[width=0.9\textwidth]{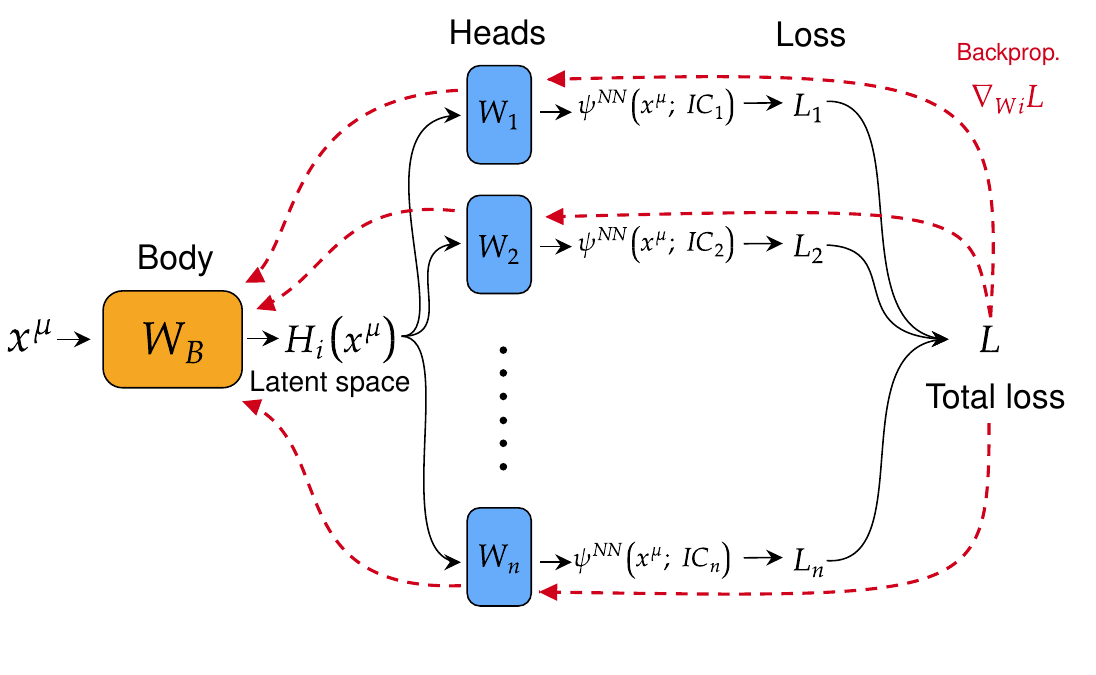}
  \caption{Graphical representation of the training graph of the MH setup. The input $x^\mu$ contains the independent variables and any variation of parameters of the DE. This is then mapped to the latent space $H_i(x^\mu)$. Then, the components of $H_i$ are linearly combined to generate the solution $\psi^{NN}$ to the DE for the $i$-th initial condition. The loss function is computed as the MSE of the residual of DE. Finally, the head weights $W_i$ and the parameters of the body $W_B$ are updated using backpropagation.}
  \label{fig: training graph}
\end{figure*}

We now describe how the model is trained. Once an approximate solution is obtained, the loss function is computed from the squared residuals of the DEs together with any additional penalty terms (Eq.~\ref{eq: loss function}). The total loss is given by the sum of the individual losses over all heads. The model is then trained by minimizing this objective. The corresponding training graph is shown in Fig.~\ref{fig: training graph}. At convergence, the model simultaneously solves all specified DE problems. Since the body is shared across all heads, it captures global properties of the solution family associated with the DE.

\subsubsection{Linear heads and head orthogonalization}\label{subsec:linear_heads_orth}

As we have mentioned, the aim of this work is to analyze the properties of the embedding space of solutions $H_i(x^\mu)$. However, it is well known that the components of the latent space are not uniquely defined due to degeneracies during the training of the model. A NN has so much freedom that, up to some accuracy given by the loss function, it can encode the same information in the latent space in many different ways. Therefore, looking at the individual components of $H_i(x^\mu)$ will not give us any valuable information, since they will strongly vary across different architectures, initialization, training schedule, etc. In this subsection we will present a method to break this degeneracy when looking at the Principal Components (PCs) of the embedding space.

We focus on the case in which each of the heads is as simple as possible: a linear layer. This simplification will allow us to gain explainability on the latent space components. In this case, the response of each of the heads can be written in the following form
\begin{equation}
    \psi^{raw}(x^\mu;IC) \,=\, \sum_{i = 1}^{n_b}w(IC)_i H^i(x^\mu)
\end{equation}
where $H^i(x^\mu)$ are the latent space components and $w(IC)$ the set of weights of the head solving the equation for a particular IC/BC. The output is then reparameterized (see Eq.~\ref{eq: parametrized NN output} below) into $\psi^{NN}(x^\mu; IC)$, such that the IC/BC is matched exactly\footnote{This is the \emph{hard} enforcement approach, in which the IC is built into the ansatz exactly.}. Note that in these expressions the weights of the head can be thought of as coefficients, and the latent space functions act as a “basis”. So, in some way, the NN is learning what is the optimal basis made out of $n_b$ functions to solve the DE problem for different ICs/BCs, for a fixed domain of the independent variables, and for a range of values of different parameters $\{x^{\mu}\}_{\mu = m,\ldots, d}$.

To make things more explicit, the response of each head can be written as follows:

\begin{equation}
\psi^{NN}(x^\mu;IC_i) = v(x;IC_i) + F(x,t)\sum_{j = 1}^{n_b}w(IC_i)_j H^j(x^\mu)
\label{eq: parametrized NN output}
\end{equation}
where $t$ represents the temporal independent variable\footnote{Our presentation focuses on the IC setting. However, the framework extends straightforwardly to BCs.}, and $v(x;IC_i)$ is the $i$-th IC at $t = 0$. Here $F(x,t)$ is an auxiliary function that allows to impose the IC at $t =0$, and the Dirichlet BC at the edges of the interval. The specific expression of this function is the following
\begin{equation*}
    F(x,t) = \frac{x -x_{min}}{x_{max} - x_{min}} \left(1 - \frac{x -x_{min}}{x_{max} - x_{min}}\right) \left(1- e^{-t}\right)
\end{equation*}
where $x_{min}$ and $x_{max}$ represent the edges of the spatial domain of interest, this is, $x \in [x_{min}, x_{max}]$. Note that the weights $w_j(IC_i)$ and the latent space components $H^j$ depend on the initialization of the NN parameters, and hence do depend on arbitrary choices that one makes. However, the full solution $\psi^{NN}(x^\mu; IC_i)$ does not depend on those choices, since it is defined as the solution to a DE with an IC.  Therefore, up to an error term of the order of the DE residual loss, we have the following relation between different latent space realizations
$$\sum_{j = 1}^{n_b} w^i_j\, H^j(x^\mu) \,=\, \sum_{j = 1}^{n_b} \hat{w}^i_j\, \hat{H}^j(x^\mu), \quad i\in\{1,\dots,N_h\},$$
where $N_h$ is the number of heads (i.e., the number of ICs/BCs solved simultaneously in the MH setup), $\hat{\,\cdot\,}$ denotes quantities obtained from an independently trained model (e.g., different random initialization), and $w^i_j \equiv w_j(IC_i)$ are the linear-head weights associated with the $i$-th IC. In matrix form, this relation reads $W H = \hat{W}\hat{H}$, where $W\in\mathbb{R}^{N_h\times n_b}$ and $H\in\mathbb{R}^{n_b}$ (with entries given by the functions $H^j$ evaluated at $x^\mu$).

In our experiments we choose $N_h=n_b$, so $W$ is a square matrix. We further assume that it is invertible. We can then left-multiply by $W^{-1}$ to obtain
\begin{equation}
H^j(x^\mu) \,=\, (W^{-1})^{j}{}_{i}\, \hat{W}^{i}{}_{k}\, \hat{H}^{k}(x^\mu).
\label{eq: transformation}
\end{equation}
In principle, we do not have to assume $W$ to be invertible. It is sufficient that $W$ has full column rank (hence $N_h\ge n_b$), so that the latent components are identifiable from the collection of heads up to an $n_b\times n_b$ linear transformation. In practice, full column rank is expected when the heads correspond to sufficiently diverse ICs/BCs and the optimization does not collapse multiple latent directions;  the head-orthogonalization procedure introduced below is designed to discourage such degeneracies. For completeness, if $N_h\neq n_b$ one may replace $W^{-1}$ by the Moore--Penrose pseudoinverse $W^{+}$ but we do not consider such a case in our work here.

We now introduce the centered latent-space components. Subtracting the mean removes the overall offset of each latent direction and isolates the fluctuations around the average latent configuration. This is the natural quantity to consider in PCA, since PCs are intended to capture directions of maximal variance rather than the mean embedding itself. Moreover, centering removes a possible rank-one contribution associated with the average latent profile, preventing the PCA spectrum from being dominated by trivial global shifts. The centered latent variables are defined as
\begin{equation*}
    H_c^{i}(x^\mu)
    \,=\,
    H^i(x^\mu)-\mu^i,
\end{equation*}
where the mean of the \(i\)-th latent component is computed over the sampled points in the \(x^\mu\) directions,
\begin{equation*}
    \mu^i
    \,=\,
    \frac{1}{M}\sum_{l=1}^{M} H^i(x^\mu_l).
\end{equation*}
where we have a total of $M$ sampling points $\{x^\mu_l\}_{l \in \{1,\dots,M\}}$ across different parameters (we use uppercase $M$ to avoid confusion with $m$, used earlier for the number of independent variables).

It is important to note that, under changes between different training realizations, the mean transforms covariantly in the same way as the latent components themselves. Indeed, since the latent transformation matrix is independent of the sampling points \(x^\mu\), one finds
\begin{equation*}
    \mu^j
    \,=\,
    (W^{-1})^{j}{}_{i}\,\hat W^{i}{}_{k}\,\hat\mu^k.
\end{equation*}
Consequently, the centered latent-space components obey the same transformation law as the original latent variables,
\begin{equation*}
    H^j_c(x^\mu)
    \,=\,
    (W^{-1})^{j}{}_{i}\,\hat W^{i}{}_{k}\,\hat H^{k}_c(x^\mu).
\end{equation*}

Since centered and non-centered latent variables transform identically, we work from this point onward with centered latent-space components and, for notational simplicity, suppress the subscript \(c\) in \(H_c^i\).

\subsection{PCA of the latent space}
Let us call the matrix $W^{-1}\hat{W} \equiv A$. In order to perform a PCA of the embedding space, we need to compute the covariance matrix. This object is defined as follows:
$$C^{ij} = \sum_{l=1}^{M} H^i(x^\mu_l) H^j(x^\mu_l).$$\footnote{We omit the conventional $1/M$ normalization for notational simplicity; including it rescales all eigenvalues by the same factor and does not affect the analysis below.}
Using the transformation rule given in Eq.~\ref{eq: transformation} we obtain that the covariance matrix transforms in the following way under latent space reparameterizations
\begin{align}
C^{ij} &= 
\sum_{l=1}^{M}\sum_{k = 1}^{n_b}\sum_{m = 1}^{n_b}
A^i_k\,\hat{H}^k(x^\mu_l)\,
A^j_m\,\hat{H}^m(x^\mu_l) \nonumber\\
&= 
\sum_{k = 1}^{n_b}\sum_{m = 1}^{n_b}
A^i_k\,\hat{C}^{km}\,(A^{\top})_m^{\,j},
\end{align}
where $\hat{C}$ corresponds to the covariance matrix in the $\hat{H}$ basis
$$\hat{C}^{ij} = \sum_{l=1}^{M} \hat{H}^i(x^\mu_l) \hat{H}^j(x^\mu_l).$$
The covariance matrix therefore transforms as
$$C = A\hat{C}A^{\top}.$$
PCA amounts to solving
\begin{equation}
C\,v = \lambda\, v,
\end{equation}
where $\lambda$ are the eigenvalues (explained variances) and $v$ are the corresponding eigenvectors (principal directions). Equivalently, the eigenvalues are the roots of the characteristic polynomial
$$\det(C-\lambda I)=0.$$

If $A\in O(n_b)$, then $C$ and $\hat{C}$ are related by an orthogonal similarity transform, $C=A\hat{C}A^{\top}$. Since $\det A \, \det A^{\top} = 1$ and $\lambda I = A(\lambda I)A^{\top}$, this gives $\det(C-\lambda I) = \det(\hat{C}-\lambda I)$, so $C$ and $\hat{C}$ share the same eigenvalues.

Hence $C$ and $\hat{C}$ have the same eigenvalues when $A\in O(n_b)$, i.e., when the change of latent basis between two trainings is orthogonal. In general, relating two latent representations requires that the head-weight matrix $W$ to be full rank (so that $W^{-1}$ exists in the square case). We enforce head orthogonalization as a practical way to (i) discourage rank collapse and (ii) make the induced change of basis approximately orthogonal, stabilizing the PCA spectrum across random initializations.

Let $W\in\mathbb{R}^{N_h\times n_b}$ denote the matrix collecting the head weights, with entries $W_{ij}=w_j(IC_i)$. In the square case used in this work ($N_h=n_b$), we add the penalty
\begin{equation}\label{eq: head orth condition}
    L_{\mathrm{add}} \,=\, \lVert WW^{\top} - I\rVert_{F}^{2} \, +\, \lVert W^{\top}W - I\rVert_{F}^{2}.
\end{equation}
 For square $W$ both terms vanish on the same set, so only one is strictly necessary; we keep both because they generate symmetric gradients and stabilize optimization. This drives $W$ to be approximately orthogonal, which in particular implies that $W$ is full rank. If both $W$ and $\hat W$ are (approximately) orthogonal, then $A=W^{-1}\hat W=W^{\top}\hat W\in O(n_b)$ (up to the accuracy enforced by the penalty), and the PCA eigenvalues become insensitive to the particular latent-space representation learned in a given training run.

A natural question at this point is whether we can quantify the error on the corresponding eigenvalues given the loss residuals. Here we provide the final result, the reader is referred to the Appendix \ref{Appendix2} for the full details. Given two NN realizations corresponding to different initial seeds, the full solutions are denoted by $\psi$ and $\hat{\psi}$, with corresponding latent space components $H$ and $\hat{H}$ as well as their weight matrices $W$ and $\hat{W}$ and their covariance matrices $C$ and $\hat{C}$. The resulting bound on the eigenvalue difference is

\[
\begin{aligned}
|\lambda_k(C)-\lambda_k(\hat C)|
&\le
\left(
\|H\|_{L^2(\mathcal D)}
+
\|A\widehat H\|_{L^2(\mathcal D)}
\right)
\\
&\quad \times
\|W^{-1}\|_2 \cdot 4L K_{H^1}
(\sqrt{\epsilon}+\sqrt{\hat{\epsilon}})
+
\delta \|\hat C\|_2
\end{aligned}
\]
with 
\begin{align*}
\epsilon := \|\mathcal R(\psi)\|^2_{L^2(\mathcal D)}
\qquad
\widehat\epsilon :=\|\mathcal R(\widehat\psi)\|^2_{L^2(\mathcal D)}
\end{align*}
and
\begin{align*}
 \delta := \|A^TA-I\|_2.
\end{align*}
This is not meant to be an optimal bound, and is only meant to illustrate the fact that one can control the error on the difference between the eigenvalues given the error on the losses as well as some regularity estimates on the approximate solutions $\psi$ and $\hat{\psi}$. In particular, when the loss residuals $\epsilon, \hat{\epsilon}$ and the orthogonalization error $\delta$ are all small, the right-hand side is small, and the PCA spectra of $C$ and $\hat{C}$ agree up to a controlled error --- so the spectrum becomes effectively independent of the training realization.

\subsection{Spectral weight decomposition}\label{subsec: spectral decomposition}

PCA orders the latent directions by explained variance, but each principal component is itself a vector in latent space and can mix contributions from many spatial scales. For a physically meaningful interpretation we would like to know \emph{which spatial frequencies are responsible for the variance captured by each PC}. The decomposition introduced in this section answers exactly that question: it splits each PC's eigenvalue $\lambda_n$ into a sum over wavenumber bands (``Fourier shells'') $Q$, giving a profile $\rho_n(Q)$ that tells us how much of $\lambda_n$ comes from each scale. This profile is invariant under the change-of-basis freedom that the orthogonalization penalty leaves in the latent space, and is therefore a training-independent observable.

The construction proceeds in three steps. (i) Fourier-transform the latent components along the spatial axis $x^1$ and group wavenumbers into shells. (ii) Define the Q-shelled covariance $C^{(Q)}$ and show it transforms in the same way as the full covariance under latent reparametrizations. (iii) Define the spectral weight $\rho_n(Q) \equiv v_n^{\top} C^{(Q)} v_n$ and verify it is independent of the training realization.

Assume we have a realization of the latent space components $H_i(x^\mu)$. We define the discrete Fourier transform with respect to the $x^1$ variable\footnote{We restrict the Fourier analysis to the spatial coordinate $x^1$, as the associated Fourier modes naturally encode the scale dependence of the solutions. However, the PC spectral decomposition introduced below is fully general and can be applied to any other coordinate or variable without modification.} as
\begin{equation*}
    \widetilde{H}^i(k, x^\nu)
    =
    \sum_{x} e^{-ikx^1} H^i(x^1, x^\nu),
\end{equation*}
where $\nu = 2,\dots,d$.

The inverse transform can be written as a sum over discrete momentum shells $Q \subset \mathcal{K}$ as
\begin{equation*}
    H^i(x^\mu)
    =
    \sum_{Q}
    \sum_{k \in Q}
    \frac{1}{2\pi}
    e^{ikx^1}
    \widetilde{H}^i(k, x^\nu),
\end{equation*}
where the sum runs over all momentum shells $\mathcal{K}$.

With this definition, the covariance matrix reads
\begin{equation*}
\begin{split}
    C_{ij}
    &=
    \sum_{l=1}^{M}
    H^i(x_l^\mu) H^j(x_l^\mu) \\
    &=
    \sum_{l=1}^{M}
    \sum_{Q,R \in \mathcal{K}}
    \sum_{k \in Q}
    \sum_{k' \in R}
    \frac{e^{ix_l^1(k+k')}}{(2\pi)^2}
    \widetilde{H}^i(k, x_l^\nu)
    \widetilde{H}^j(k', x_l^\nu)\\
    &=
    \sum_{l=1}^{M}
    \sum_{Q}
    \sum_{k \in Q}
    \frac{1}{2\pi}
    \widetilde{H}^i(k, x_l^\nu)
    \widetilde{H}^j(-k, x_l^\nu) \\
    &=
    \sum_Q C^{(Q)}_{ij},
\end{split}
\end{equation*}
where we have defined the Q-shelled covariance matrix as
\begin{equation*}
    C^{(Q)}_{ij}
    \equiv
    \sum_{l=1}^{M}
    \sum_{k \in Q}
    \frac{1}{2\pi}
    \widetilde{H}^i(k, x_l^\nu)
    \widetilde{H}^j(-k, x_l^\nu).
\end{equation*}
Note that to simplify the expression of the covariance matrix we have used the discrete Fourier identity
\begin{equation*}
    \sum_{x_l} e^{i x_l (k+k')}
    =
    2\pi \delta_{k,-k'}.
\end{equation*}

It is important to note that the Q-shelled covariance matrix transforms in the same way as the full covariance matrix. Recall that the Fourier transformed latent space components transform as
\begin{equation*}
    \widetilde{H}^i(k,x^\nu)
    =
    A_{ij}\widetilde{\hat{H^{\,j}}}(k,x^\nu),
\end{equation*}
where the $\hat{\cdot}$ denotes quantities obtained from an independent training run. The matrix $A$ is defined in terms of the head weights as $A \equiv W^{-1}\hat{W}$.

From this we obtain the transformation of the Q-shelled covariance matrix,
\begin{equation*}
\begin{split}
    C_{ij}^{(Q)}
    &=
    \sum_{l=1}^{M}
    \sum_{k \in Q}
    \frac{1}{2\pi}
    A_{im}\widetilde{\hat{H}^{\,m}}(k, x_l^\nu)
    A_{jn}\widetilde{\hat{H}^{\,n}}(-k, x_l^\nu) \\
    &=
    A_{im}A_{jn}
    \hat{C}^{(Q)}_{mn}.
\end{split}
\end{equation*}
In matrix notation,
\begin{equation}\label{eq: Q-shelled rule}
    C^{(Q)} = A \hat{C}^{(Q)} A^{\top}.
\end{equation}
This is the same transformation rule as for the full covariance matrix. It follows immediately that
\begin{equation*}
    \mathrm{spec}(C^{(Q)}) = \mathrm{spec}(\hat{C}^{(Q)}),
\end{equation*}
provided that $A$ is orthogonal as is the case in our usual setup. Hence both Q-shelled covariance matrices share the same eigenvalues, denoted $\{\lambda_n^{(Q)}\}$.

The Q-shelled covariance matrix admits a natural spectral interpretation. In particular, its trace measures the total power carried by the Fourier modes contained within the momentum shell $Q$,
\begin{equation*}
\begin{split}
P(Q)
=
\sum_{k\in Q}\sum_i
\widetilde{H}^{i}(k,x^\nu)\,
\widetilde{H}^{i}(-k,x^\nu)= \mathrm{Tr}\left(C^{(Q)}\right).
\end{split}
\end{equation*}
Thus, the trace of $C^{(Q)}$ can be interpreted as the contribution of the shell $Q$ to the total variance of the latent representation. Moreover, this quantity is invariant under latent-space reparametrizations, as follows directly from the transformation rule of the Q-shelled covariance matrix given in Eq.~\ref{eq: Q-shelled rule}.

In general, the eigenvalues of $C^{(Q)}$ do not sum to those of the full covariance matrix,
\begin{equation*}
    \lambda_n \neq \sum_Q \lambda_n^{(Q)},
\end{equation*}
unless the commutativity condition holds,
\begin{equation*}
    [C^{(Q)}, C^{(R)}] = 0, \quad \forall Q,R
\end{equation*}
which is generally not the case.

However, one can define the Q-shelled spectral weight as
\begin{equation*}
    \rho_n(Q)
    \equiv
    v_n^T C^{(Q)} v_n,
\end{equation*}
where $v_n$ is the eigenvector of the full covariance, satisfying $C v_n = \lambda_n v_n$.

This quantity measures the contribution of shell $Q$ to the total variance captured by the $n$-th principal component. Summing over all shells one recovers the full eigenvalue,
\begin{equation*}
    \lambda_n
    =
    \sum_Q v_n^T C^{(Q)} v_n
    =
    \sum_Q \rho_n(Q).
\end{equation*}

Finally, we verify that this quantity is invariant under latent space reparameterizations. Using that the eigenvector transforms as $v_n = A\hat{v}_n$, with
\begin{equation*}
    \hat{C}\hat{v}_n = \lambda_n \hat{v}_n
    \quad \Rightarrow \quad
    C(A\hat{v}_n) = \lambda_n (A\hat{v}_n),
\end{equation*}
we compute
\begin{equation*}
\begin{split}
    \rho_n(Q)
    &=
    (A\hat{v}_n)^{\top} C^{(Q)} (A\hat{v}_n) \\
    &=
    \hat{v}_n^{\top} A^{\top} C^{(Q)} A \hat{v}_n \\
    &=
    \hat{v}_n^{\top} A^{\top} A \hat{C}^{(Q)} A^{\top} A \hat{v}_n \\
    &=
    \hat{v}_n^T \hat{C}^{(Q)} \hat{v}_n
    =
    \hat{\rho}_n(Q),
\end{split}
\end{equation*}
where we assumed that the eigenvalues are non-degenerate so that eigenvectors transform as $v_n = A\hat{v}_n$.

The PCA spectrum of the latent space admits a fully invariant decomposition into Fourier shells, where each eigenvalue $\lambda_n$ is expressed as a sum of scale-resolved contributions $\rho_n(Q)$. In this sense, PCs are not tied to a single characteristic frequency, but instead define robust spectral profiles over momentum space that are stable under latent reparametrizations induced by different training realizations. In Sec.~\ref{sec: results}, we apply this framework to the viscous Burgers, heat and wave equations, and show that the dominant spectral shells for each principal component cluster around intermediate scales, with a dependence on the IC family.

\subsection{Initial conditions and model specifics}\label{subsect ICs}
A key element in our pipeline is the choice of ICs, which are used to probe the structure of the solution manifold learned by the model. We distinguish three complementary families of ICs: Fourier modes, polynomials, and wavelets. In this work, we always train the models with $N_h = n_b = 20$ ICs drawn from these families. We also impose vanishing Dirichlet boundary conditions at the edges of the spatial domain for all the families, i.e., $\psi^{NN}(t, x_{min}; IC) = \psi^{NN}(t, x_{max}; IC) = 0$.

Rather than being arbitrary choices, these IC families are selected to probe qualitatively different aspects of the solution space geometry. Fourier ICs provide globally supported, spectrally pure perturbations and therefore probe the model's ability to represent and disentangle individual frequency components. Polynomial ICs emphasize smooth, low-frequency variations and test the capacity of the representation to capture globally coherent trends and non-oscillatory structure. In contrast, wavelet ICs are spatially localized and inherently multiscale, allowing us to probe how the learned manifold encodes localized structures, sharp features, and scale-dependent interactions. Together, these three families form a minimal but complementary basis of probes that span global-to-local and smooth-to-oscillatory regimes.

The explicit construction of these ICs, including the precise sampling procedures for the Fourier coefficients, polynomial bases, and wavelet parametrizations, is deferred to Appendix~\ref{appendix: IC}, as these details are implementation-specific and not essential for understanding the geometric role of the IC families within our framework.

\section{Results}
\label{sec: results}

\begin{table*}[t]
\centering
\setlength{\tabcolsep}{12pt}
\renewcommand{\arraystretch}{1.35}
\resizebox{\textwidth}{!}{
\begin{tabular}{l l c c c c c}
\toprule
\textbf{Equation} & \textbf{Initial condition} & \textbf{Min. Loss} & \textbf{PC (90\%)} & \textbf{PC (95\%)} & \textbf{PC (99\%)} & \textbf{Dom. shells (PC1-PC3)} \\
\midrule
\multirow{3}{*}{\textit{Burgers}} & Fourier & $2.06 \times 10^{-3}$ & 3 & 4 & 6 & $Q_{3}$, $Q_{2}$, $Q_{4}$ \\
& Polynomials & $4.15 \times 10^{-4}$ & 2 & 2 & 4 & $Q_{3}$, $Q_{2}$, $Q_{3}$ \\
& Wavelets & $6.77 \times 10^{-3}$ & 3 & 4 & 7 & $Q_{3}$, $Q_{4}$, $Q_{4}$ \\
\cmidrule(lr){2-7}
\multirow{3}{*}{\textit{Heat}} & Fourier & $1.16 \times 10^{-4}$ & 3 & 3 & 5 & $Q_{4}$, $Q_{2}$, $Q_{3}$ \\
& Polynomials & $2.52 \times 10^{-4}$ & 1 & 1 & 2 & $Q_{2}$, $Q_{1}$, $Q_{1}$ \\
& Wavelets & $1.53 \times 10^{-2}$ & 3 & 4 & 6 & $Q_{3}$, $Q_{4}$, $Q_{4}$ \\
\cmidrule(lr){2-7}
\multirow{3}{*}{\textit{Wave}} & Fourier & $1.67 \times 10^{-2}$ & 4 & 5 & 7 & $Q_{4}$, $Q_{2}$, $Q_{3}$ \\
& Polynomials & $4.72 \times 10^{-2}$ & 2 & 2 & 3 & $Q_{2}$, $Q_{1}$, $Q_{2}$ \\
& Wavelets & $7.62 \times 10^{-2}$ & 3 & 4 & 6 & $Q_{3}$, $Q_{4}$, $Q_{4}$ \\
\bottomrule
\end{tabular}}
\caption{Minimum loss function and number of principal components required to recover 90\%, 95\%, and 99\% of the latent-space variance, for each equation and initial condition basis. The total dimension of the latent space is $20$. Dominant shells are listed compactly as the top dyadic shell for each PC from PC1 to PC3.}
\label{tab:latent_pca_summary}
\end{table*}

\begin{figure*}[t]
    \centering
    \includegraphics[width=\textwidth]{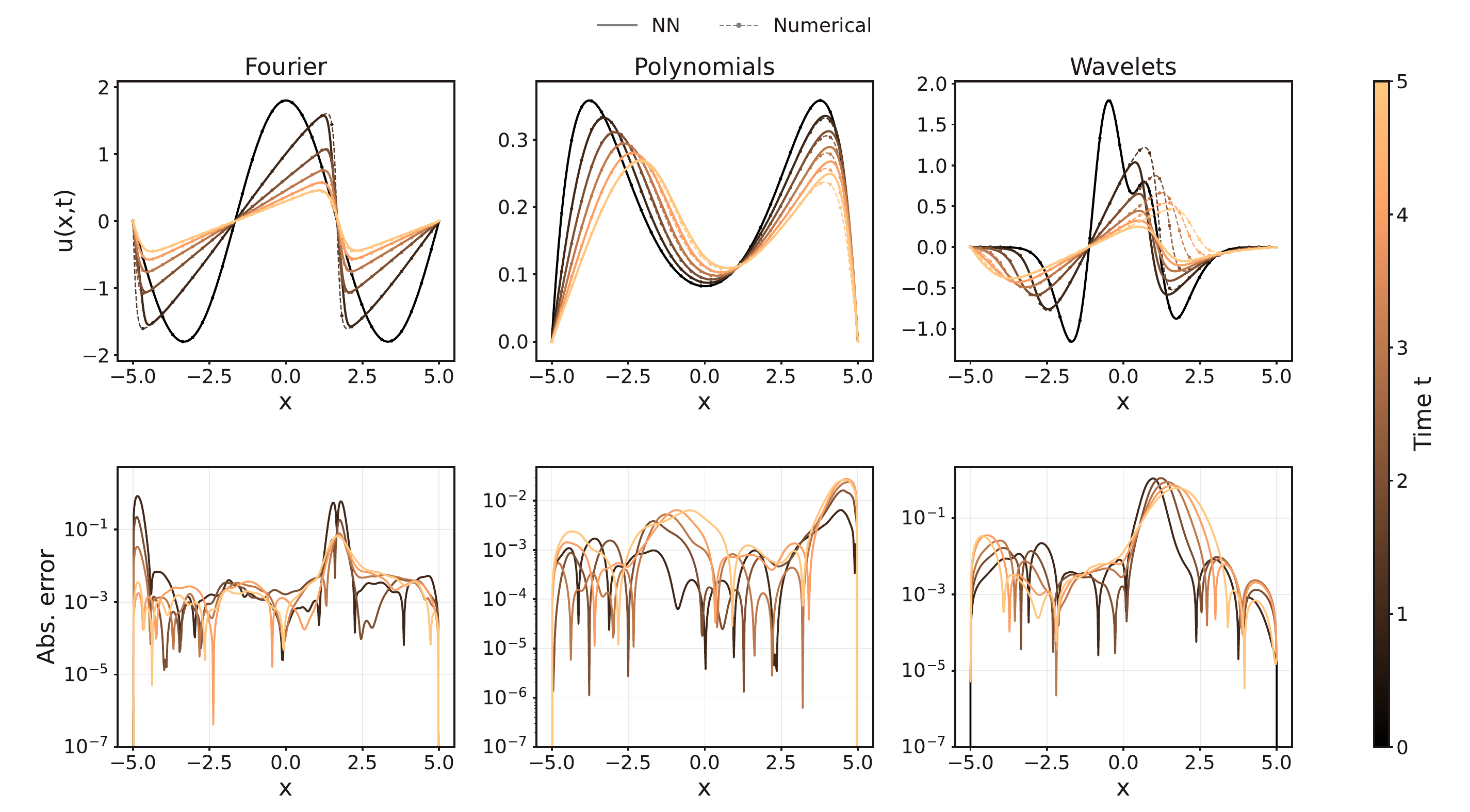}
    \caption{Comparison between numerical solutions (dashed lines) and multihead PINN predictions (solid lines) for the one-dimensional Burgers equation with viscosity $\nu=0.1$. Results are shown for representative initial conditions drawn from the three families considered in this work: Fourier modes (left), polynomials (center), and wavelets (right). Different colors correspond to different time snapshots of the evolution. The lower panels display the absolute error between the neural-network prediction and the numerical solution.}
    \label{fig:burgers_solutions}
\end{figure*}

In this section, we present results for the one-dimensional Burgers equation (Eq.~\ref{eq: burgers equation}) using three families of ICs: Fourier modes, polynomials, and wavelets.

\begin{figure*}[t]
    \centering
    \includegraphics[width=1.0\linewidth]{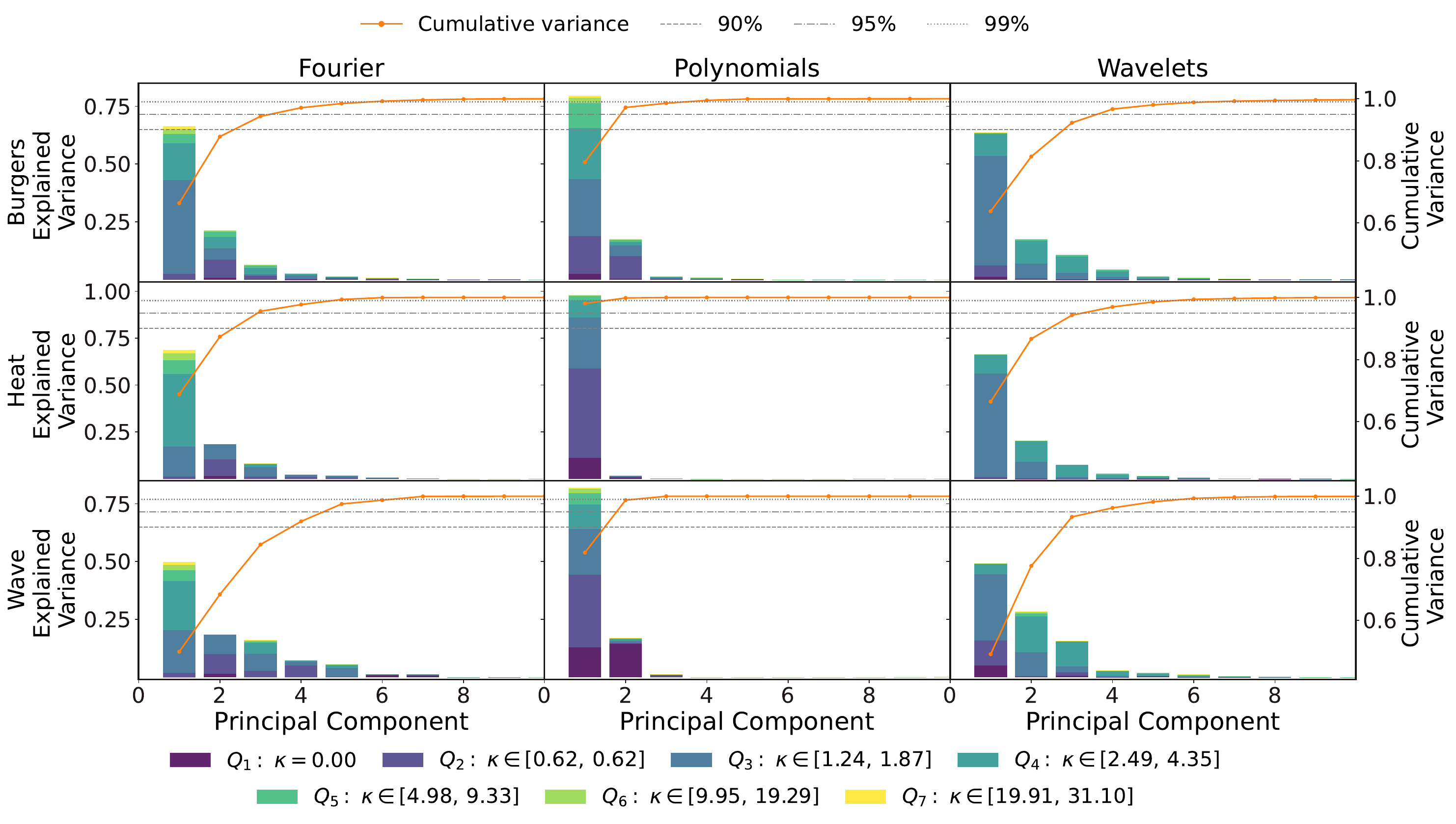}
    \caption{Principal-component spectra of the learned latent space for all equations and initial-condition families. Each bar shows the explained variance ratio $\lambda_n / \sum_m \lambda_m$ for each principal component, with colors indicating the contribution of each momentum shell $Q$ according to the spectral weight decomposition $\rho_n(Q)/\sum_m\lambda_m$ (Subsec.~\ref{subsec: spectral decomposition}). The orange curve and right axis show the cumulative explained variance; dashed lines mark the 90\%, 95\%, and 99\% thresholds. Rows correspond to Burgers (top), heat (middle), and wave (bottom); columns to Fourier, polynomial, and wavelet initial-condition families.}
    \label{fig:pca_grid}
\end{figure*}

\begin{figure*}[t]
    \centering
    \includegraphics[width=1.0\linewidth]{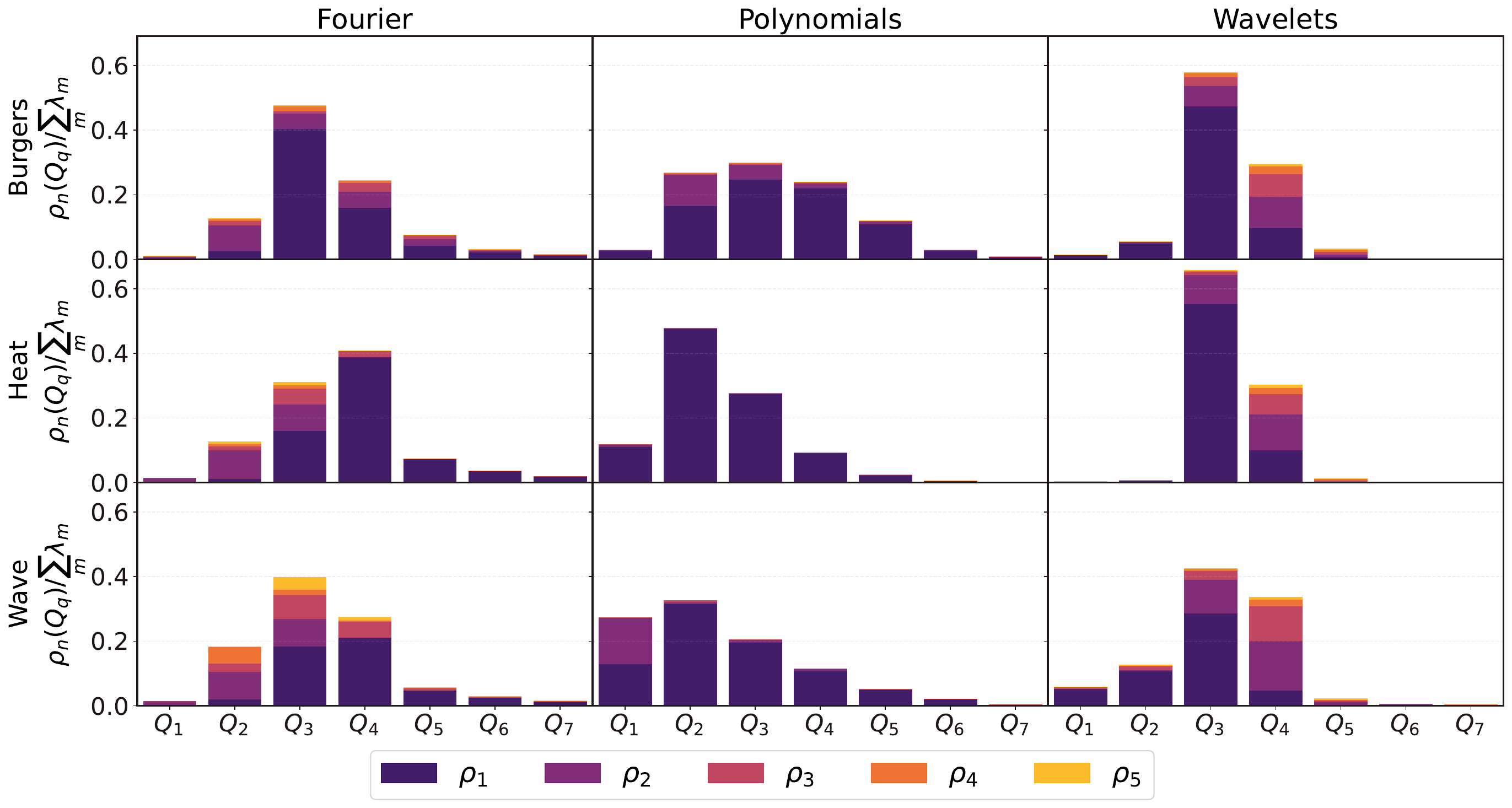}
    \caption{Spectral weight decomposition $\rho_n(Q)/\sum_m \lambda_m$ as a function of momentum shell $Q$ for the leading five principal components ($\rho_1$ through $\rho_5$), shown for each combination of equation (rows: Burgers, heat, wave) and initial-condition family (columns: Fourier, polynomials, wavelets). The stacked bars indicate the contribution of each principal component to the total latent-space variance carried by shell $Q$, taking into account just the five most relevant principal components.}
    \label{fig:shells_hist}
\end{figure*}

The main quantitative result is that the learned embeddings have a much smaller effective dimension than the nominal latent dimension $n_b=20$. Across the Burgers experiments, $2$--$4$ principal components are sufficient to recover $95\%$ of the latent-space variance, while $4$--$7$ components recover $99\%$, depending on the IC family. The same compression pattern persists for the heat and wave equations, indicating that the effect is not specific to a single PDE or to a particular choice of ICs (Fig. \ref{fig:pca_grid}). We therefore use the PCA spectrum as the primary diagnostic of effective dimensionality, and complement it with the Fourier-shell decomposition introduced in Sec.~\ref{subsec: spectral decomposition} with dyadic shells, which assigns scale-resolved weights to the leading principal components (Figs. \ref{fig:pca_grid}, \ref{fig:shells_hist}). This lets us check whether each leading PC corresponds to a distinct spatial scale, or instead mixes contributions across many scales. Our results are made degeneracy-free by the head-orthogonalization penalty (Eq. \ref{eq: head orth condition}). 

In order to complete our analysis and further demonstrate the robustness of our results, we also applied the method to the heat and wave equations. From a mathematical perspective, both the heat and Burgers equations are dissipative systems. In particular, the heat equation is linear and admits an explicit solution in Fourier space. The evolution is obtained by acting with the heat kernel on the initial condition, leading to an exponential suppression of the Fourier modes,
\begin{equation}
\hat{u}(k,t)=e^{-\kappa k^2 t}\hat{u}_0(k).
\end{equation}
This behavior is qualitatively similar to the decay observed in the Principal Component Analysis (PCA) spectrum (Fig. \ref{fig:pca_grid}). To further assess the robustness of our conclusions, we also considered a non-dissipative linear PDE, namely the wave equation, whose Fourier modes evolve as
\begin{equation}
\hat{u}(k,t)=A_k e^{ivkt}+B_k e^{-ivkt},
\end{equation}
and therefore exhibit purely oscillatory, rather than dissipative, behavior.

A summary of the results is provided in Table~\ref{tab:latent_pca_summary}. Details of the NN architecture, the collocation points, and the training procedure are provided in Appendix~\ref{appendix: NN}. A comparison between the numerical solutions and those obtained with the PINN is shown in Figs.~\ref{fig:burgers_solutions}, \ref{fig: solutions_heat}, and \ref{fig: solutions_wave}, for the Burgers, heat, and wave equations, respectively. We observe good agreement between the numerical and PINN solutions, with absolute errors of order $10^{-2}$. The worst-performing case corresponds to the wavelet IC family for the Burgers equation. We attribute the mismatch occurring near $x \simeq 2$ to the formation of a near-shock structure, which makes it difficult for the PINN to converge in that region.

A crucial point is that the PCA is not performed on the full solution
\(u(x,t)\). The IC is imposed explicitly through the
ansatz\footnote{For wave equation we slightly change the parametrization of the solution to
$$
F(x,t) = \frac{x -x_{min}}{x_{max} - x_{min}} \left(1 - \frac{x -x_{min}}{x_{max} - x_{min}}\right) \left(1- e^{-t^2}\right)
$$
This choice automatically enforces the IC
$$
\partial_t \psi^{\rm NN}(x,0;\mathrm{IC}) = 0,
$$
since the prefactor $(1-e^{-t^2})$ and its first time derivative both vanish at $t=0$.}
\begin{equation}
\psi^{\rm NN}(x,t;\mathrm{IC})
=
v(x;\mathrm{IC})
+
F(x,t)
\sum_{j=1}^{n_b} w_j(\mathrm{IC})H_j(x^\mu).
\end{equation}
Thus the latent functions \(H_j\) represent the residual degrees of
freedom required to evolve away from the prescribed IC.
The PCA spectrum should therefore be interpreted as the effective
dimension of this residual solution manifold, rather than of the full
solution family including the initial data.

This distinction is especially relevant for the wave equation. A PCA of
the full solution family can be dominated by the imposed IC. For instance, for orthogonal Fourier ICs in
the wave equation, the full solutions remain approximately orthogonal in
space-time, leading to a nearly flat spectrum at sufficiently long times.
Our analysis removes this trivial contribution and instead probes the
remaining latent degrees of freedom learned by the shared body of the
network. Further discussion regarding this point is included in the wave equation appendix \ref{Appendix 1}.

The results in both Table~\ref{tab:latent_pca_summary} and Fig. \ref{fig:pca_grid} show that all three
equations exhibit a pronounced reduction in effective latent dimension, but with clear dependence on both the PDE and the IC family. Polynomial ICs lead to the strongest compression, while Fourier and wavelet ICs generally require more PCs. The wave equation with Fourier IC is the least compressed case, consistent with the absence of diffusive damping.
Thus, the PCA spectrum should be viewed as a property of the equation-constrained residual manifold sampled by a chosen ensemble of ICs, not as an intrinsic dimension of the entire PDE solution space.

We next use the Fourier-shell decomposition to assign a scale-resolved
profile to each leading PC (Figs. \ref{fig:pca_grid} and \ref{fig:shells_hist}). For the \(n\)-th principal
component \(v_n\), the decomposition gives
\begin{equation}
\lambda_n = \sum_Q \rho_n(Q),
\qquad
\rho_n(Q)=v_n^{\top} C^{(Q)}v_n ,
\end{equation}
where \(C^{(Q)}\) is the contribution of Fourier shell \(Q\) to the
latent covariance. Since \(C^{(Q)}\) transforms covariantly under the
same latent reparametrizations as the full covariance, the weights
\(\rho_n(Q)\) provide degeneracy-free spectral profiles for the PCA
modes.

The shell profiles should be interpreted conservatively. They do not
define universal preferred frequencies for Burgers, heat, or wave
dynamics. The dominant shells depend visibly on the IC
family. What can be read directly from
Table~\ref{tab:latent_pca_summary} and Figs. \ref{fig:pca_grid}, \ref{fig:shells_hist} is that, for the first three PCs, the
largest shell contribution lies in \(Q_1\)--\(Q_4\) across the
experiments considered. This suggests that, in the present setup, the
dominant residual variance is associated with low-to-intermediate
spatial scales. The precise shell locations, however, remain conditioned
by the chosen IC family and by the shell discretization.

The physical role of the shell decomposition is therefore not to identify
a universal frequency band, but to show that the degeneracy-free PCA
modes have a scale-resolved structure. For heat, the residual latent
degrees of freedom encode diffusion-induced changes of the initial
profile. For Burgers, they encode the deformation generated by nonlinear
advection together with viscous damping. For wave dynamics, the absence
of diffusion is reflected most clearly in the weaker PCA compression for
Fourier ICs. In all cases, the Fourier-shell analysis
turns the abstract PCA hierarchy into a set of scale-resolved observables
of the learned residual manifold.

\section{Conclusions}
\label{sec: conclusions}

In this work we have studied the structure of embedding (or latent) spaces associated with neural-network representations of families of PDE solutions. Although our numerical analysis uses the one-dimensional viscous Burgers equation as a controlled testbed, the central point is more general: a latent space learned from a family of PDE solutions can be regarded as a geometric object carrying information about the corresponding solution manifold. The present study should therefore be viewed as a first step toward extracting interpretable geometrical information from latent spaces of NNs trained on PDEs in general.

Using a MH PINN architecture, we have learned a shared latent representation encoding families of solutions with different ICs and parameter values. The shared body defines the embedding space, while the solution-specific heads reconstruct individual members of the family. This formulation shifts the focus from approximating isolated solutions to characterizing the geometry of the solution space itself. While tested here for 1D problems, the construction is in principle applicable to any PDE family amenable to PINN training, though convergence in higher-dimensional systems remains to be demonstrated.

A central result is that the learned latent space exhibits pronounced low-dimensional organization. With linear heads and orthogonality constraints, the representation admits a stable PC decomposition with rapidly saturating explained-variance spectra across the tested families of ICs. For Burgers dynamics, only 2--4 PCs are sufficient to capture approximately 95\% of the latent-space variance, despite a nominal embedding dimension of 20. This hierarchy is not imposed by construction, but emerges from the simultaneous enforcement of the governing equations and the representation of a family of solutions. A complementary result is provided by the Fourier-shell decomposition of the latent covariance: the spectral weights $\rho_n(Q)$ reveal that the dominant PCs concentrate their variance at low-to-intermediate momentum shells ($Q_1$--$Q_4$), with the precise distribution depending on the IC family. This scale-resolved structure is invariant under admissible latent reparametrizations and therefore constitutes a robust, training-independent observable of the learned embedding. The dependence of both the effective dimensionality and the shell-resolved spectral profiles on the equation and on the IC ensemble suggests that the latent representation reflects properties of the chosen solution family, rather than merely providing a generic low-dimensional compression.

We further find that the effective dimensionality and spectral organization of the latent space depend on the family of ICs used during training. Polynomial ICs, which are spectrally concentrated at low wavenumbers, require fewer PCs and shift the dominant spectral weight toward lower shells compared to Fourier or wavelet families. This IC dependence is a feature of the method rather than a limitation: it indicates that the learned manifold faithfully reflects the structure of the solution family it was trained on, rather than producing a universal representation independent of the input ensemble.

More broadly, the latent hierarchy suggests that neural representations of PDEs may encode intrinsic collective degrees of freedom. The ordering of modes is defined by explained variance in latent space rather than by spatial wavelength, providing a data-driven form of scale organization. This bears a structural similarity to Wilsonian (\cite{wilson1974renormalization}) coarse-graining: a high-dimensional description of a
family of solutions is mapped onto a hierarchy of progressively more relevant latent degrees of freedom, while less important directions contribute increasingly little to the representation. The Fourier-shell decomposition further suggests an interpretation in terms of scale-dependent organization.
Although no explicit renormalization-group transformation or flow is constructed here, the observed hierarchy indicates that latent spaces may provide a natural framework for identifying effective variables in nonlinear PDEs. Establishing whether a genuine RG structure underlies these learned
representations remains an interesting direction for future work \citep{chen1996renormalization, Forster1977, Goldenfeld2018}.

This perspective raises several structural questions for future work. One natural direction is to define a metric tensor on the learned manifold, which would provide a coordinate-independent characterization of its geometry and help distinguish robust physical structure from artifacts of parametrization or 
optimization. Similarly, approximate equivariances associated with symmetries of the underlying PDE---such as translations, rotations, scalings, Galilean transformations, or conservation laws---may define natural flows in latent space and provide a principled link between representation learning and 
coarse-graining.

The implications extend beyond Burgers equation. Neural networks trained on PDEs may provide access not only to approximate solutions, but also to the geometry of the full solution space. This is particularly relevant for nonlinear and multiscale systems, where high-dimensional simulations often obscure the 
underlying organization. Potential applications include Navier--Stokes dynamics, reaction--diffusion systems, kinetic equations, plasma models, relativistic evolution equations, and PDEs arising in astrophysics and cosmology. In cosmological and astrophysical contexts, large-scale structure formation, 
turbulence, and other nonlinear systems involve coherent structures, multiscale interactions, and expensive high-dimensional simulations \citep{Angulo2012, Villaescusa-Navarro2021}. Operator-learning and 
effective-field-theory approaches already indicate the existence of reduced descriptions \citep{Li2021FNO, Lu2021DeepONet}. Our results suggest a complementary interpretation: such reductions may be understood geometrically as properties of learned latent manifolds.

We emphasize that the present work is exploratory. While the analysis in this paper is restricted to one-dimensional spatial systems, the underlying method is not inherently limited to 1D, although the computational cost is substantial (approximately 10 days per experiment on an NVIDIA H100). The learned representation is also specific to the IC family used during training rather than universal. Extensions to higher-dimensional systems, forced dynamics, non-dissipative equations, complex boundary conditions, and turbulent regimes are required to assess the generality of the observed structure. Future work will apply this framework to the Navier--Stokes equations and other nonlinear PDEs, develop coordinate-independent diagnostics of latent geometry, and test whether similar hierarchies persist in more realistic physical settings. In this broader sense, NNs may serve not only as solvers or emulators, but also as tools for uncovering the geometry of PDE solution spaces.

\bmhead{Acknowledgements}

We thank Pablo Tejerina-P\'erez, Pau Sol\'e Vilar\'o, Marisol Traforetti, and Ali Kalout for useful discussions on PINNs, multihead architectures, and model explainability.

This work was supported by a grant from the Simons Foundation (00017375, RJ).
Funding for the work of RJ, LS and PT was partially provided by project PID2022-141125NB-I00,
and the “Center of Excellence Maria de Maeztu 2025-2029” award to the ICCUB funded by
grant CEX2024-001451-M from AEI/10.13039/501100011033.
PT is supported by the project “Dark Energy and the Origin of the Universe” (PRE2022-102220), funded by MCIN/AEI/10.13039/501100011033.

\bmhead{AI disclaimer and LLM usage}
The AI multi-agent \href{https://astropilot-ai.github.io/DenarioPaperPage/}{\em Denario} suggested exploring principal-component analyses of neural-network embeddings as a possible avenue for studying the structure of latent spaces (see pp.~23--26 of \cite{denario}). The present work develops this initial suggestion into a complete research program, including the design of the multihead PINN framework, the construction of latent-space observables, the numerical experiments, the analysis of the results, and the interpretation of the emerging geometric structure.

Standard AI-assisted programming tools were used during software development to improve efficiency in routine tasks such as debugging, code refactoring, and modification of existing implementations. Large language models were also used to assist with language editing and manuscript polishing. All scientific decisions, methodological choices, numerical experiments, analyses, and conclusions were carried out by the authors.

\bmhead{Code availability}
The code used to generate the results is available at the \href{https://github.com/pedrota2000/PDE_embeddings}{GitHub repository}. It is built on a modified version of the existing \texttt{Neurodiffeq} package (\cite{chen2020neurodiffeq, liu2025recent}), a Python library developed for the implementation of Physics-Informed Neural Networks (PINNs) in \texttt{PyTorch}.

\begin{appendices}

\section{Neural network architecture and training details}\label{appendix: NN}

We summarize in this appendix the technical details of the neural network architecture and the training setup.

The three tests performed for the Burgers equation have the same technical specifications. The temporal independent variable is chosen to be in the interval $t\in [0,5]$, whereas the spatial variable is taken to be $x \in [-5,5]$. We sample both variables from an equally spaced distribution, with $100$ points on each direction. As we have already mentioned, we are interested in solving for different values of the viscosity parameter $\nu$. We solve for $25$ values of this parameter sampled from a logarithmic distribution from $10^{-2}$ to 1. For the \textit{body} model we have used a FCNN composed of $5$ layers of $128$ neurons each. Following Sec.~\ref{subsec:linear_heads_orth}, we set the latent-space dimension to $n_b=20$ and train $N_h=20$ linear heads, each associated with a different initial condition. The head is chosen to be a linear layer without activation or bias. The model is then trained for $3.0 \cdot 10^{5}$ epochs. The loss function is chosen to be the residual of DE squared, multiplied by a gradient annihilation term that allows the model to better resolve the points at which the derivatives are large (\cite{FerrerSanchez2024}). An additional term that forces the head orthogonalization technique is added to the loss function, as shown in Eq.~\ref{eq: head orth condition}. The hyperparameter chosen to multiply this additional loss was found experimentally, and it has a value of $\lambda = 1\cdot 10^{-3}$. The optimization algorithm used is Adam, with initial learning rate $10^{-3}$ and the default parameters given by \texttt{pytorch}. An initial linear warm-up phase is done during $1000$ epochs from a learning rate of $10^{-6}$ to facilitate the convergence of the model. During training a sequential learning rate scheduler has been used. This scheduler reduces the learning rate by a factor of $0.985$ every $1000$ epochs. These models have been trained on an NVIDIA H100 GPU. Each test took $\sim 10$ days of computational time. Despite the large training cost, this investment yields a foundational model that captures a broad class of solutions within a single framework. Beyond predictive accuracy, the learned latent space provides access to meaningful geometric and dynamical information, offering a degree of interpretability that is typically absent in solution-specific models.

For the heat equation, the technical specifications of the model are the same as the ones used for Burgers equation. The only difference is that we no longer use the gradient annihilation term, since the heat equation is not expected to develop steep gradients as the thermal diffusivity goes to zero. In this case, the viscosity $\nu$, is replaced by the thermal diffusivity $\kappa$. The training domain for the independent variables $x$ and $t$ is exactly the same. The range of $\kappa$ is the same as the one for the viscosity $\nu$.

For the wave equation the specifications of the setup are similar to the ones we used for Burgers equation. However, there are two main differences. The first one is that we again do not use a gradient annihilation term, since the wave equation is not expected to develop steep gradients. The second one is that this equation is no longer exponentially sensitive to the value of the parameter $v$. It is for this reason that we choose to use a set of $25$ points linearly sampled between $0.1$ and $2$. The sampling of the independent variable is identical to the one used in the previous cases.

\section{Initial condition generation}\label{appendix: IC}
In this appendix, we explain how the ICs for the different families introduced in the main text are generated. On top of these ICs, we impose vanishing Dirichlet BCs at the edges of the spatial interval.

\subsection{Fourier components}
We start by explaining how ICs for the Fourier family are generated. As we have explained in Subsec.~\ref{subsect ICs}, the model is trained for $N_h = 20$ different ICs. For this particular family of IC, we generate $10$ sines and $10$ cosine components that vanish on the edges of the $x$ interval specified in the previous appendix. To be concrete, the particular expression for the $i$-th IC can be written as follows
\begin{equation}
    v_{\text{F}}(x; IC_i) \,=\, \left\{
    \begin{split}
        A_{s,i} \sin\left(\frac{x n_i \pi}{5}\right), \qquad \text{if} \quad i\leq 10\\
        B_{s,i}\cos\left(\frac{xm_i \pi}{10}\right), \qquad \text{if} \quad i>10
    \end{split}
    \right.
\end{equation}
where $A_{s,i}$ and $B_{s,i}$ are random numbers sampled for each IC from a uniform distribution between $0.1$ and $2.0$. $n_i$ is a random integer between $1$ and $5$, and $m_i$ a random odd integer between $1$ and $5$.

\subsection{Polynomials}
In this subsection we explain how different ICs for the polynomial case are generated. We have trained the model with $N_h = 20$ different ICs. Each of those has been generated as the sum of three different polynomials with degree in between $2$ and $7$, all of them vanishing at the edges of the interval. These three polynomials have been combined with coefficients in between $0$ and $1$, sampled from a uniform random distribution. More concretely, these ICs can be expressed as follows
\begin{equation}
    v_{\text{P}}(x; IC_i) \,=\,\sum_j  C^i_{j}  \,  p^{a_j}(x)
\end{equation}
where $C^{i}_j$ are coefficients sampled from a random uniform distribution, $a_j$ is a random integer number in between $2$ and $7$, and $p^{i}$ denotes a polynomial of degree $i$ that vanishes at the edges of the interval.

\subsection{Wavelets}
We explain how different ICs are sampled in the wavelet case. We have trained the model with $N_h = 20$ different ICs. These are generated by linearly combining $10$ Ricker wavelets with randomly generated scales and shifts. More concretely, the ICs for this case are generated using the following expression
\begin{equation}\label{eq: wvlts}
    v_W(x; IC_i) \,=\, \sum_{j = 1}^{10} C^{i}_j  \, \Psi\left(\frac{x-B^i_j}{S^i_j}\right)
\end{equation}
where $\Psi$ is the Ricker wavelet defined as follows:
\begin{equation*}
    \Psi(x) \,=\, \frac{2}{\sqrt{3} \pi^{1/4}} \left(1-x^2 \right) e^{-\frac{x^2}{2}}.
\end{equation*}
$C^{i}_j$ are coefficients sampled from a uniform random distribution between $0$ and $1$. The shift ($B^i_j$) and the scale ($S^i_j$) of the wavelet are sampled from a uniform random distribution between $-1$ and $1$, and $0.45$ and $1.0$ respectively.

\section{Heat and Wave equations}\label{Appendix 1}
In this appendix, we present both the heat and the wave equations. The PCA decomposition of the latent spaces, and their corresponding spectral weight decomposition are shown in Figs. \ref{fig:pca_grid} and \ref{fig:shells_hist}. 

\subsection{Heat equation}

\begin{figure*}
    \centering
    \includegraphics[width=1.0\linewidth]{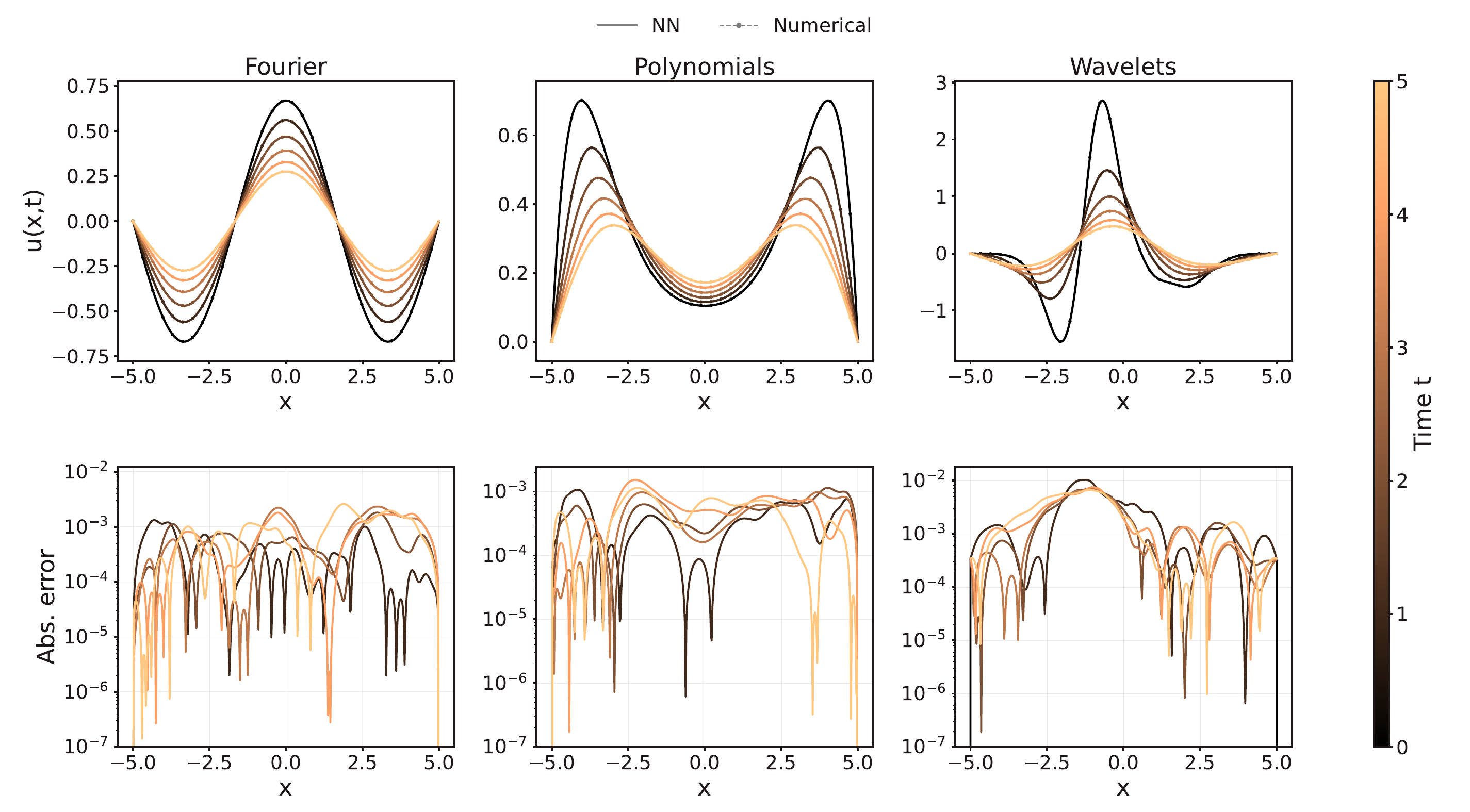}
    \caption{Solutions of the one-dimensional heat equation reconstructed by the multihead PINN for a thermal diffusivity $\kappa=0.2$. Columns correspond to representative Fourier, polynomial, and wavelet initial conditions. Solid and dashed lines denote the neural-network prediction and numerical solution, respectively, while colors indicate different times during the evolution. The lower panels show the corresponding absolute errors.}
    \label{fig: solutions_heat}
\end{figure*}

The heat equation is a second-order, linear, partial differential equation that is used to describe the evolution of a temperature profile on a domain of interest. This equation can be written in $1$ spatial dimension in the usual following form
\begin{equation*}
    \frac{\partial u(x,t)}{\partial t} \,=\, \kappa \frac{\partial^2u(x,t)}{\partial x^2}
\end{equation*}
where $u(x,t)$ represents the temperature profile, and $\kappa$ is the thermal diffusivity.

We train the model for different sets of ICs. We generate them in the same way as we have explained for Burgers equation. We consider the same three cases: fourier components, polynomials and wavelets. We show the corresponding results in Figs. \ref{fig:pca_grid}, \ref{fig:shells_hist} and \ref{fig: solutions_heat}.

\subsection{Wave equation}

\begin{figure*}
        \centering
    \includegraphics[width=1.0\linewidth]{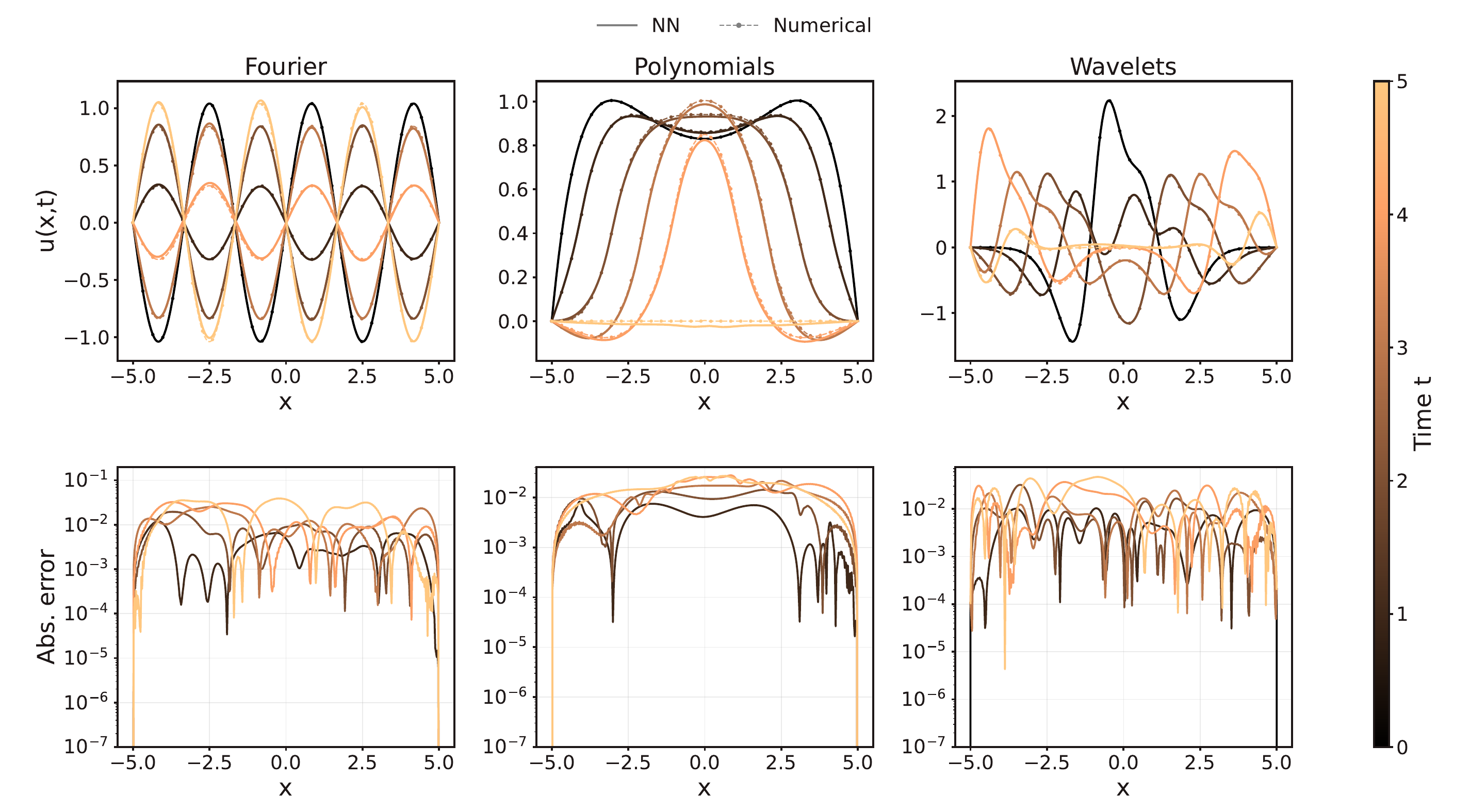}
    \caption{Solutions of the one-dimensional wave equation reconstructed by the multihead PINN for a wave velocity $v=1$. Columns correspond to representative Fourier, polynomial, and wavelet initial conditions. Solid and dashed lines denote the neural-network prediction and numerical solution, respectively, while colors indicate different times during the evolution. The lower panels display the absolute error between the neural-network prediction and the numerical solution.}
    \label{fig: solutions_wave}
\end{figure*}

The wave equation is a second order, non-dissipative, linear, partial differential equation. In one spatial dimension, this equation can be written in the following form
\begin{equation*}
    \frac{\partial^2 u(x,t)}{\partial t^2} \,=\, v^2 \frac{\partial^2 u(x,t)}{\partial x^2}
\end{equation*}
where $u$ is a scalar quantity that propagates as a wave, and $v$ is a positive real number that represents the propagation speed of the wave. 

Note that since wave equation is second order in time, we also need to fix the partial derivative with respect to time at the initial time step. We do so by slightly changing the $F(x,t)$ function in the reparameterization of the NN to
\begin{equation*}
    F(x,t) = \frac{x -x_{min}}{x_{max} - x_{min}} \left(1 - \frac{x -x_{min}}{x_{max} - x_{min}}\right) \left(1- e^{-t^2}\right)
\end{equation*}

We train the model for three different sets of ICs. The way of generating these initial conditions is the same as the one we used for both the Burgers and the heat equation. We show the results obtained for the wave equation in Figs. \ref{fig:pca_grid}, \ref{fig:shells_hist} and \ref{fig: solutions_wave}.

We now discuss how our results are related to doing the PCA directly on the solution for the Fourier IC family. Intuitively one might expect the PCA to look flat for orthonormal Fourier modes. This is only true if we do the PCA on the full solution set not on the latent space elements. Let us sketch the argument. Suppose we have a set of $k=1,..,K$ orthonormal fourier modes 

$$\langle \phi_k,\phi_{k'}\rangle_x = \delta_{kk'}$$

which satisfy the eigenvalue equation $\partial_{xx}\phi_k = -\lambda_k\phi_k$ (for instance $\lambda_k = (2\pi k/L)^2$ for a domain of length $L$ with periodic boundary conditions). The solution to the wave equation with initial conditions $u_k(x,0)=\phi_k(x)$, $\partial_t u(x,0)=0$, is easily seen to be $u(x,t)=\phi_k(x)\cos(v\sqrt{\lambda_k}t)$. Then doing the PCA on the full solution, 

\begin{align*}
\langle u_k,u_{k'} \rangle_{x,t}
&= \delta_{kk'} \int_0^T \cos^2(v\sqrt{\lambda_k}t)\,dt \\
&= \frac12 \delta_{kk'}
\left(
T + \frac{\sin(2v\sqrt{\lambda_k}T)}{2v\sqrt{\lambda_k}}
\right).
\end{align*}
At large $T$, the first term dominates, so the spectrum is flat and there is no dimensionality reduction.

\section{Quantitative Eigenvalue Estimate Error}
\label{Appendix2}

Here we will prove the bounds for the continuum limit case, i.e. where sums are integrals etc. We define the following $L^2$ norm

$$\|H\|_{L^2(\mathcal D)} := \sqrt{\int_{\mathcal D}\sum_i (H^i(\xi))^2}$$

Sometimes we will need to estimate the norm, say, only in the spatial variable $x$ then the obvious definition is 

$$\|H(t,\nu)\|_{L_x^2} := \sqrt{\int\sum_i (H^i(x,t,\nu))^2 dx}$$

Suppose we have two initial seed runs corresponding to $H^i$ and $\hat{H}^i$, and for notational convenience most of the times we will drop the independent variable arguments of $H^i$ and other associated quantities. Then we can write their corresponding parametrizations as 

\begin{align*}
\psi^i - v^i = \alpha(t) g(x) W^i_j H^j\\
\hat{\psi}^i - v^i = \alpha(t)g(x) \hat{W}^i_j \hat{H}^j
\end{align*}

Where $\alpha(t) = 1-e^{-t}$. And $g(x)$ is defined by

\begin{equation*}
    g(x) = \frac{x -x_{min}}{x_{max} - x_{min}} \left(1 - \frac{x -x_{min}}{x_{max} - x_{min}}\right) 
\end{equation*}

Hence to compare the eigenvalues of the two corresponding covariance matrices, we first take the difference between the two equations:

\begin{equation*}
    W^i_j H^j - \hat{W}^i_j\hat{H}^j = \frac{\psi^i - \hat{\psi}^i}{(1-e^{-t})g(x)} 
\end{equation*}

Now multiplying by $W^{-1}$ we get:

\begin{equation*}
        H^i - (W^{-1}\hat{W})^i{}_j\hat{H}^j = (W^{-1})^i{}_j\frac{\psi^j - \hat{\psi}^j}{(1-e^{-t})g(x)}
\end{equation*}

Then it follows that:

\begin{equation*}
     \|{H - (W^{-1}\hat{W})\hat{H}}\|_{L^2(\mathcal D)} \le \|(W^{-1})\|_2 \|Q_F(\psi-\hat\psi)\|_{L^2(\mathcal D)} 
\end{equation*}

Where $Q_F(\psi-\hat\psi)=\frac{\psi - \hat{\psi}}{(1-e^{-t})g(x)}$ and the norms $\|.\|_F$ and $\|.\|_2$ refer to the Frobenius and operator norms respectively.

Now define the empirical latent covariance matrices

\begin{equation*}
C_{ij} := \langle H^i,H^j\rangle_{L^2(\mathcal D)} 
\qquad
\hat C_{ij} := \langle \hat{H}^i,\hat{H}^j\rangle_{L^2(\mathcal D)} 
\end{equation*}

We compare $C$ not directly to $\hat C$, but to the transformed covariance. In the continuum setting,

\begin{align*}
A\widehat C A^T
&=
\int_{\mathcal D}
(A\widehat H)(\xi)(A\widehat H)(\xi)^T\,d\xi .
\end{align*}

Set

\begin{align*}
X(\xi):=H(\xi),
\qquad
Y(\xi):=A\widehat H(\xi).
\end{align*}

Then

\begin{align*}
C-A\widehat C A^T
&=
\int_{\mathcal D}
\left[
X(\xi)X(\xi)^T
-
Y(\xi)Y(\xi)^T
\right]\,d\xi .
\end{align*}

Using

\begin{align*}
X(\xi)X(\xi)^T-Y(\xi)Y(\xi)^T
&=
(X(\xi)-Y(\xi))X(\xi)^T
+\\
\quad Y(\xi)(X(\xi)-Y(\xi))^T
\end{align*}

Then

\begin{align*}
C-A\widehat C A^T
&=
\int_{\mathcal D}
\left[ X(\xi)X(\xi)^T -Y(\xi)Y(\xi)^T
\right]\,d\xi .
\end{align*}

Using
\[
\begin{aligned}
X(\xi)X(\xi)^T-Y(\xi)Y(\xi)^T
&=
(X(\xi)-Y(\xi))X(\xi)^T
\\
&\quad
+
Y(\xi)(X(\xi)-Y(\xi))^T .
\end{aligned}
\]

we get

\noindent
\(\displaystyle
\begin{aligned}
\|C-A\widehat C A^T\|_2
&\le
\int_{\mathcal D}
\left\|
X(\xi)X(\xi)^T
-
Y(\xi)Y(\xi)^T
\right\|_2\,d\xi
\\[0.4em]
&\le
\int_{\mathcal D}
\Big(
\|(X(\xi)-Y(\xi))X(\xi)^T\|_2
\\
&\qquad\qquad
+
\|Y(\xi)(X(\xi)-Y(\xi))^T\|_2
\Big)d\xi
\\[0.4em]
&\le
\int_{\mathcal D}
\Big(
|X(\xi)-Y(\xi)|_2\,|X(\xi)|_2
\\
&\qquad\qquad
+
|Y(\xi)|_2\,|X(\xi)-Y(\xi)|_2
\Big)d\xi
\\[0.4em]
&\le
\|X-Y\|_{L^2(\mathcal D)}
\|X\|_{L^2(\mathcal D)}
\\
&\qquad
+
\|Y\|_{L^2(\mathcal D)}
\|X-Y\|_{L^2(\mathcal D)}
\\[0.4em]
&=
\left(
\|X\|_{L^2(\mathcal D)}
+
\|Y\|_{L^2(\mathcal D)}
\right)
\|X-Y\|_{L^2(\mathcal D)} 
\end{aligned}
\)

Therefore,

\begin{align*}
&\|C-A\widehat C A^T\|_2\\
&\le
\left(
\|H\|_{L^2(\mathcal D)}
+
\|A\widehat H\|_{L^2(\mathcal D)}
\right)
\|H-A\widehat H\|_{L^2(\mathcal D)}
\end{align*}

Combining this with the previous estimate gives:

\[
\begin{aligned}
\|C-A\hat C A^T\|_2
&\leq
\left(
\|H\|_{L^2(\mathcal D)}
+
\|A\widehat H\|_{L^2(\mathcal D)}
\right)
\|W^{-1}\|_2
\\
&\quad \times
\|Q_F(\psi-\hat\psi)\|_{L^2(\mathcal D)} 
\end{aligned}
\]

We now compare the eigenvalues of $C$ and $\hat C$. Since $A$ is not necessarily orthogonal, $A\hat C A^T$ is not exactly orthogonally similar to $\hat C$. Therefore we write

\begin{align*}
|\lambda_k(C)-\lambda_k(\hat C)|
&\leq
|\lambda_k(C)-\lambda_k(A\hat C A^T)|
\\
&\quad
+
|\lambda_k(A\hat C A^T)-\lambda_k(\hat C)| .
\end{align*}

Since $\hat{C}$ is symmetric positive semidefinite it has a square root, and hence we can write it as 

$$\hat{C}=\hat{C}^{\frac12} \hat{C}^{\frac12}$$

The non zero eigenvalues of $A \hat{C} A^T$ are the same as that of $\hat{C}^{\frac12} A^T A\hat{C}^{\frac12}$, hence by Weyl's eigenvalue theorem we have 

$$|\lambda_k(A \hat{C} A^T) - \lambda_k(\hat{C})|\le \|\hat{C}^{\frac12} A^T A\hat{C}^{\frac12} -\hat{C}\|_2$$

Hence we can write

\begin{align*}
|\lambda_k(A \hat{C} A^T) - \lambda_k(\hat{C})|
&\le
\|\hat{C}^{1/2} (A^T A - I)\hat{C}^{1/2}\|_2
\\
&\le
\|\hat{C}\|_2 \, \|A^T A - I\|_2 
\\
&:= \delta \|\hat{C}\|_2 
\end{align*}

where we defined $\delta = \|A^TA-I\|_2$ i.e. the orthogonalization error. Using Weyl's eigenvalue theorem one more time, we get

\begin{align*}
&|\lambda_k(C)-\lambda_k(A\hat C A^T)| \le \|C-A\hat{C}A^T\|_2 \\
&\leq \left(
\|H\|_{L^2(\mathcal D)} +
\|A\widehat H\|_{L^2(\mathcal D)}
\right)
\|W^{-1}\|_2 \, \|Q_F(\psi-\hat\psi)\|_{L^2(\mathcal D)}
\end{align*}

We now estimate the quotient appearing in the latent-space perturbation.
Let
\begin{equation*}
z:=\psi-\hat\psi,
\end{equation*}
and define the Burgers residuals
\begin{equation*}
R(\psi):=\psi_t+\psi\psi_x-\nu\psi_{xx},
\qquad
R(\hat\psi):=\hat\psi_t+\hat\psi\hat\psi_x-\nu\hat\psi_{xx}.
\end{equation*}
We denote
\begin{equation*}
\epsilon:=\|R(\psi)\|_{L^2(D)}^2,
\qquad
\hat\epsilon:=\|R(\hat\psi)\|_{L^2(D)}^2.
\end{equation*}

Since the hard-enforcement factor is
\begin{equation*}
F(x,t)=g(x)(1-e^{-t}),
\end{equation*}
with
\begin{equation*}
g(x)=
\frac{x-x_{\min}}{L}
\left(
1-\frac{x-x_{\min}}{L}
\right),
\qquad
L=x_{\max}-x_{\min},
\end{equation*}
the quotient appearing in the latent-space perturbation is
\begin{equation*}
Q_F z
:=
\frac{z}{g(x)(1-e^{-t})}.
\end{equation*}

Let
\begin{equation*}
d(x,\partial\Omega)
=
\min\{|x-x_{\min}|,\ |x_{\max}-x|\}.
\end{equation*}
Then
\begin{equation*}
g(x)
=
\frac{(x-x_{\min})(x_{\max}-x)}{L^2}
=
\frac{d(x,\partial\Omega)\bigl(L-d(x,\partial\Omega)\bigr)}{L^2}.
\end{equation*}
Since $d(x,\partial\Omega)\leq L/2$, we have
\begin{equation*}
g(x)\geq \frac{d(x,\partial\Omega)}{2L},
\end{equation*}
and therefore
\begin{equation*}
\frac{1}{g(x)}
\leq
\frac{2L}{d(x,\partial\Omega)}.
\end{equation*}

Both realizations satisfy the same homogeneous Dirichlet boundary conditions
and the same initial condition. Hence
\begin{equation*}
z(x_{\min},t)=z(x_{\max},t)=0,
\qquad
z(x,0)=0.
\end{equation*}
Thus the numerator vanishes on the same parts of the space-time boundary
on which the hard-enforcement factor vanishes. To control the corresponding
quotient on the full spatial domain, we use the one-dimensional Hardy
inequality
\begin{equation*}
\left\|
\frac{u}{d(x,\partial\Omega)}
\right\|_{L^2(\Omega)}
\leq
2\|u_x\|_{L^2(\Omega)},
\end{equation*}
valid for functions $u$ vanishing on $\partial\Omega$. Taking
\begin{equation*}
u
=
\frac{z}{1-e^{-t}},
\end{equation*}
we obtain
\begin{equation*}
\begin{split}
\|Q_F z\|_{L^2(D)}
&=
\left\|
\frac{z}{g(x)(1-e^{-t})}
\right\|_{L^2(D)} \\
&\quad \leq
4L
\left\|
\partial_x
\left(
\frac{z}{1-e^{-t}}
\right)
\right\|_{L^2(D)} .
\end{split}
\end{equation*}

We use the following $H^1_x$ stability estimate:
\begin{equation*}
\left\|
\partial_x
\left(
\frac{z}{1-e^{-t}}
\right)
\right\|_{L^2(D)}
\leq
K_{H^1}
\left(
\sqrt{\epsilon}+\sqrt{\hat\epsilon}
\right),
\end{equation*}
where $K_{H^1}$ depends on the domain, final time, viscosity, and regularity
bounds on $\psi$ and $\hat\psi$. It follows that
\begin{equation*}
\|Q_F z\|_{L^2(D)}
\leq
4L K_{H^1}
\left(
\sqrt{\epsilon}+\sqrt{\hat\epsilon}
\right).
\end{equation*}

Hence, putting everything together we have:

\[
\begin{aligned}
|\lambda_k(C)-\lambda_k(\hat C)|
&\le
\left(
\|H\|_{L^2(\mathcal D)}
+
\|A\widehat H\|_{L^2(\mathcal D)}
\right)
\\
&\quad \times
\|W^{-1}\|_2 \cdot 4L K_{H^1}
(\sqrt{\epsilon}+\sqrt{\hat{\epsilon}})
+
\delta \|\hat C\|_2
\end{aligned}
\]

\end{appendices}


\end{document}